\documentclass[sigconf]{acmart}

\usepackage{algorithm}
\usepackage{algpseudocode}
\usepackage{makecell}   
\usepackage{multirow}
\usepackage{tikz}
\usepackage{threeparttable}
\usepackage{tabularx}
\newenvironment{DIFnomarkup}{}{}

\renewcommand{\footnotetextcopyrightpermission}[1]{}
\newcommand*{\emptycirc}[1][1ex]{\tikz\draw (0,0) circle (#1);}
\newcommand*{\halfcirc}[1][1ex]{%
\begin{tikzpicture}
\draw[fill] (0,0) -- (90:#1) arc (90:270:#1) -- cycle;
\draw (0,0) circle (#1);
\end{tikzpicture}}
\newcommand*{\fullcirc}[1][1ex]{\tikz\fill (0,0) circle (#1);}
\AtBeginDocument{%
  }
\AtBeginDocument{%
  }

\setcopyright{acmlicensed}
\copyrightyear{2026}
\acmYear{2026}
\acmDOI{XXXXXXX.XXXXXXX}

\acmISBN{978-1-4503-XXXX-X/2018/06}

\begin{document}

\title{EdgeRefine: Privacy-Utility Balance for Graphs via Jaccard Sampling under Edge Differential Privacy}

\author{Wenxiu Ding}
\authornote{Correspondence to wxding@xidian.edu.cn}
\affiliation{%
  \institution{State Key Laboratory of Integrated Services Networks, School of Cyber Engineering, Xidian University}
  \city{Xi'an}
  \state{Shaanxi}
  \country{China}
}

\author{Muzhi Liu}
\affiliation{%
  \institution{State Key Laboratory of Integrated Services Networks, School of Cyber Engineering, Xidian University}
  \city{Xi'an}
  \state{Shaanxi}
  \country{China}
  }

\author{Zheng Yan}
\affiliation{%
  \institution{State Key Laboratory of Integrated Services Networks, School of Cyber Engineering, Xidian University}
  \city{Xi'an}
  \state{Shaanxi}
  \country{China}
}

\author{Mingjun Wang}
\affiliation{%
  \institution{School of Cyber Engineering, Xidian University}
  \city{Xi'an}
  \state{Shaanxi}
  \country{China}
  }

\author{Yifan Zhao}
\affiliation{%
  \institution{School of Cyber Engineering, Xidian University}
  \city{Xi'an}
  \state{Shaanxi}
  \country{China}
  }

\author{Qiao Liu}
\affiliation{%
  \institution{School of Cyber Engineering, Xidian University}
  \city{Xi'an}
  \state{Shaanxi}
  \country{China}
  }

\begin{abstract}
   Graph Neural Networks (GNNs) have shown considerable success in learning from graph-structured data. However, their application in privacy-sensitive areas remains difficult as the structural information of graphs is prone to leaking sensitive link information. To satisfy edge-level differential privacy, a common approach is to directly inject  noise into all elements of the graph's adjacency matrix, thereby obfuscating the existence of any single edge. While increased noise strengthens privacy protection, excessive noise reduces utility. Privacy-utility balance becomes a major barrier to practical privacy-preserving graph learning.

To address this issue, we propose EdgeRefine, a new local differential privacy framework that rethinks the privacy-utility trade-off in graph learning through adaptive edge refinement. 
EdgeRefine first estimates edge-existence probabilities using Jaccard similarity and ranks edges accordingly for noisy edge removal.
To ensure the sparsity and reliability of the final graph, we leverage the privacy budget $\epsilon$ to determine the ratio of true to false edges, sample them separately based on the previously obtained probability ranking, and then control the total number of edges with a separate sampling rate $k$. We conducted extensive experiments to evaluate the effectiveness of EdgeRefine, which achieves accuracy comparable to the noise-free baseline and performs much better than other privacy-preserving methods on various datasets and GNN architectures. Under privacy budget $\epsilon=2.5$, EdgeRefine achieves significant node classification accuracy improvements over state-of-the-art baselines: 17.8\% on ACM under GAT and 19.7\% on Cora under GCN. In  graph classification task, an average accuracy degradation of around 5\% has also been achieved compared to noise-free baseline. Under graph reconstruction attacks, EdgeRefine maintains relative absolute error levels consistently above 1 across all privacy budgets (averaging 1.962 on Cora and 1.472 on AMAP), indicating strong resilience against privacy leakage. 

\end{abstract}



\maketitle

\section{Introduction}
\label{intro}
Graph Neural Networks (GNNs)~\cite{GNN} are powerful methods for learning from graph-structured data~\cite{graphdata}. They have shown great success in many areas, such as social network analysis~\cite{social}, recommendation systems~\cite{rec1}, and bioinformatics~\cite{bio}. GNNs use message-passing mechanisms. By iteratively aggregating information from neighboring nodes, GNNs capture complex patterns and learn representations directly from graph structures. This very capability, while driving their rapid adoption in privacy-sensitive domains like social networks, also introduces significant privacy risks~\cite{secsurvey}. 

Differential Privacy (DP)~\cite{dp} enables the injection of noise into the original raw data, which provides a potential approach to graph data analysis with strong privacy enhancement. Common DP methods include the Laplace mechanism for continuous data ~\cite{dp} and the randomized response for binary data ~\cite{rr}. These mechanisms used in graph learning can ensure that sensitive connections are not leaked. The application of DP thereby establishes a foundation for practical privacy-preserving graph analysis in various privacy-sensitive scenarios.

 Despite these advances, current differentially private graph learning methods face fundamental limitations in balancing privacy and utility, as detailed in Table~\ref{comp}. They suffer from a number of problems, as described below:

\begin{table*}[t]
\centering
\caption{Comparison of Privacy-Preserving Graph Learning Methods}
\label{comp}

\setlength{\tabcolsep}{1pt} % 减少列间距

\begin{tabular*}{\linewidth}{@{\hspace{2pt}} @{\extracolsep{\fill}}c|ccccccc@{}@{\hspace{3pt}} }
\toprule
\textbf{Method} & 
\makecell{Pri-PGD\cite{pripgd}} & 
\makecell{PrivGraph\cite{privgraph}} & 
\makecell{LDPGen\cite{ldpgen}} & 
\makecell{LapGraph\cite{lapgraph}} & 
\makecell{Solitude\cite{solitude}} & 
\makecell{Blink\cite{blink}} & 
\makecell{\textbf{EdgeRefine}} \\
\midrule
Differential Privacy & \emptycirc[0.8ex] & \fullcirc[0.8ex] & \fullcirc[0.8ex] & \fullcirc[0.8ex] & \fullcirc[0.8ex] & \fullcirc[0.8ex] & \fullcirc[0.8ex] \\
Sparsity Preservation & \halfcirc[0.8ex] & \halfcirc[0.8ex] & \emptycirc[0.8ex] & \halfcirc[0.8ex] & \fullcirc[0.8ex] & \halfcirc[0.8ex] & \fullcirc[0.8ex] \\
Downstream Task Support & \fullcirc[0.8ex] & \emptycirc[0.8ex] & \fullcirc[0.8ex] & \fullcirc[0.8ex] & \fullcirc[0.8ex] & \fullcirc[0.8ex] & \fullcirc[0.8ex] \\
Utility-Privacy Balance & \emptycirc[0.8ex] & \halfcirc[0.8ex] & \emptycirc[0.8ex] & \emptycirc[0.8ex] & \emptycirc[0.8ex] & \halfcirc[0.8ex] & \fullcirc[0.8ex] \\
\bottomrule
\end{tabular*}

\begin{tablenotes}
  \centering
  \small
  \item[] \fullcirc[0.8ex]: Fully satisfied;
  \halfcirc[0.8ex]: Partially satisfied;
  \emptycirc[0.8ex]: Not satisfied or not involved.
\end{tablenotes}

\label{tab:comparison}
\end{table*}

(i) Existing approach fails to identify enough trustworthy edges in the noise graph adjacency matrix. 
Trustworthy edges are defined as those candidate edges that rank highest in terms of computed existence probability in the noisy graph after perturbation and similarity assessment. Identifying these edges from noisy images can better support downstream tasks, but existing work~\cite{pripgd, ppdp, sgnn} sacrifices some privacy to ensure better recognition. This makes it impossible for them to strictly guarantee differential privacy. Meanwhile, differentially private graph publication schemes ~\cite{privgraph, privdpr, psgraph} preserve high-confidence edges based on structural similarity. These methods ensure a low privacy budget, partially achieve a balance between privacy and utility, but have not yet been tested on downstream tasks such as node classification. This problem affects the practical implementation of privacy-preserving graph learning systems.

(ii) The method of denoising through probability estimation does not have a good strategy for filtering noisy edges. Although it is theoretically possible to restore the true existence of edges based on probability, the probability values estimated based on noisy data inevitably deviate, making it difficult to convert the probability matrix into a deterministic graph structure. Despite aiming to preserve the number of original edges, the sampling strategy in~\cite{GraphPub} often fails to achieve a faithful reconstruction of the overall graph structure. This is because its sampling setting is based on the assumption that probability estimation is very accurate and that estimation based on noisy data inevitably results in errors. Blink~\cite{blink} considers edges with estimated probabilities greater than a threshold as existing. However, this strategy may result in too many false edges being retained, as the overall probability of existence may be overestimated. This also degrades classification performance under certain privacy budgets, breaking the balance between privacy and utility.

(iii) The noise introduced by existing methods affects the structural rationality of the graph. Noise added for privacy damages graph sparsity which reduces the accuracy in downstream tasks. To solve this issue, Solitude framework~\cite{solitude} adds norm constraints to control the graph sparsity, however, it ignores the original properties of the graph. As a result, it performs poorly under low privacy budgets. 
To mitigate the structural distortion caused by noise, DPRR~\cite{dprr} perturbs the degree sequence with noise and then reconstructs it, whereas LAPGRAPH~\cite{lapgraph} retains the top-k noisy edges.
However, these approaches still do not preserve graph sparsity well, because this still introduces considerable noise, making the graph denser. Therefore, they are unable to maintain usable performance under low privacy budgets.

Given these problems, we need a graph learning framework that not only provides strong privacy protection but also maintains practical value for real applications. The main challenge is to maintain the utility of GNNs without breaking strict privacy guarantees.

\textbf{Technical Challenges:} However, solving these problems faces several key difficulties:

\textbf{TC1}: Estimating edge existence probabilities accurately is difficult due to the substantial noise required by a low privacy budget. Differential privacy requires adding noise based on global sensitivity. This creates a basic conflict between privacy and accuracy. For graph data, high sensitivity means that we need to add more noise. This makes it hard to obtain correct probability values. Common estimation methods often do not work well in this situation.

\textbf{TC2}: The strategy of converting probability into a deterministic structure is difficult to choose. Sampling method must ensure the use of trustworthy edges and also consider fault tolerance for errors caused by probability estimation. Common approach~\cite{blink} often produces graphs that are not accurate. Mistakes in estimation combined with random sampling cause graphs to be different each time. This leads to unstable performance in important tasks like node classification.

\textbf{TC3}: Keeping graphs sparse while satisfying privacy requirements is challenging. The accuracy of graph learning tasks, such as node classification in noisy graphs, is affected by sparsity. Because real-world networks are usually sparse, privacy protection requirements can lead to a large amount of noise being added. This mismatch causes too many artificial edges to be added. The graph becomes denser than it should be, which affects both the computational efficiency and the actual meaning of the graph structure.

\textbf{Our Proposal.} We propose the EdgeRefine framework to denoise the adjacency matrix of noisy graphs and obtain optimized graphs. The client side adds noise to the graph adjacency matrix using strict differential privacy standards, while the server side obtains an optimized graph through a two-stage denoising process for downstream tasks such as node classification. In the first stage, edge existence probabilities are estimated via Jaccard similarity computation and quantile-based binning, which groups edges with similar structural patterns to improve probability calibration under differential privacy constraints. The second stage determines the optimal ratio of selected real edges to noisy fake edges using the privacy budget $\epsilon$ and controls graph sparsity by applying a global sampling rate $k$. We utilize the privacy budget to calculate the theoretical true edge ratio $\rho = e^{\epsilon}/(1+e^{\epsilon})$ , thereby enabling us to filter out true edges and recover erroneously deleted edges respectively. The sampling rate $k$ limits the total number of selected edges to $K = \lfloor k \times E \rfloor$, where $E$ is the original edge count, the lower sampling ratio ensures fault tolerance for probability estimation bias. This integrated approach addresses three key technical challenges: \textbf{TC1} is addressed through variance-reduced probability estimation that leverages local graph topology; \textbf{TC2} is tackled by deterministically balancing real and fake edge selection based on the estimated probabilities; and \textbf{TC3} is resolved by explicitly constraining the sampling rate to preserve graph sparsity while providing formal differential privacy guarantees.

\textbf{Evaluation.}  Extensive experimental evaluations demonstrate the effectiveness of our proposed EdgeRefine framework across multiple benchmark datasets, GNN architectures, and privacy settings. We conducted comprehensive node classification experiments on four real-world graph datasets (ACM, DBLP, AMAP, and Cora) using three representative GNN architectures (GAT, GCN, and GIN) under varying privacy budgets. The results show that EdgeRefine maintains accuracy levels closely approaching the noise-free baseline while providing formal privacy guarantees. Notably, EdgeRefine exhibits exceptional stability with significantly lower variance ($\sigma^2$ = 0.0001--0.0530) and coefficient of variation (CV = 0.01--0.36) in node classification accuracy compared to competing approaches. Parameter analysis reveals an optimal sampling rate that balances edge credibility and graph connectivity, achieving peak performance at intermediate sampling values. Ablation studies confirm the critical importance of each component in our framework. In terms of node classification accuracy, the complete system outperforms the ablated version by an average of 124.6\% across all experimental conditions. Consistently, EdgeRefine also achieves stable performance on graph classification tasks. Beyond the aforementioned evaluations, we conducted additional adversarial attack experiments, further demonstrating its strong privacy-preserving capability. These comprehensive experiments validate EdgeRefine's ability to effectively address the fundamental challenges in privacy-preserving graph learning.

\textbf{Contribution.} The main contributions of this work are summarized as follows:

\begin{itemize}
    \item  We develop a novel framework, EdgeRefine, that maintains analytical utility while providing strong privacy guarantees through deterministic edge refinement. This framework effectively balances privacy constraints and practical application needs in graph neural networks.

    \item  We propose an edge existence probability estimation method based on Jaccard similarity and histogram binning. This method accurately estimates edge probabilities from noisy graph data. 

    \item  We design a sampling approach with precise rate control that eliminates selection variance. The method maintains optimal privacy-utility trade-offs through parameterized sampling rates.

    \item  We provide rigorous theoretical analysis to demonstrate that EdgeRefine is computationally efficient while meeting strict privacy requirements, accompanied by extensive experiments on real-world graph datasets. The experimental results show that EdgeRefine achieves superior performance over state-of-the-art baselines in terms of accuracy, confirming its practical effectiveness.
\end{itemize}

\section{Related Work}
\label{relatedwork}
Existing research on graph differential privacy has primarily advanced along two lines: one focused on applying formal privacy models to the graph data for analysis, and the other on developing privacy-protecting techniques within GNNs. 

\subsection{Edge-Level Differential Privacy}

Edge-level differential privacy~\cite{edgedp} is a specific privacy model designed for graph data, which focuses on protecting the existence of any single edge in the graph. It ensures that an adversary cannot reliably infer whether a particular connection between two nodes exists, even with access to the rest of the graph and auxiliary information. Research in this domain primarily focuses on developing mechanisms that enable the release of useful graph data or support training models on graph data while offering formal guarantees against edge inference attacks~\cite{graphdp}. However, the relationships between entities (edges) are interdependent, and modifying one edge can affect the connectivity  of neighboring nodes~\cite{dpsurvey, dpsurvey2}. 

Several methods have been proposed to achieve edge-level DP. For instance, PrivGraph~\cite{privgraph} leverages the community structure of graphs: it first partitions nodes into communities in a differentially-private manner and then reconstructs edges based on noisy counts of intra-community and inter-community connections, aiming to preserve global graph statistics. However, its utility has been primarily evaluated on retaining global graph statistics rather than on the performance of downstream graph learning tasks like node classification with GNNs. Another approach, LAPGRAPH~\cite{lapgraph}, perturbs the adjacency matrix directly using the Laplace mechanism and then applies sparsification to retain a subset of noisy edges for GNN training. While effective against certain attacks, its utility can be limited. Other frameworks like LDPGen~\cite{ldpgen}, which operates under the local differential privacy model for decentralized graph generation, and DPRR~\cite{dprr}, which combines randomized response with graph degree preservation, further explore the trade-offs between privacy and utility in graph data release. A common challenge among these methods is preserving sufficient data utility for complex machine learning tasks, particularly those involving GNNs.

\subsection{Privacy-Preserving Graph Neural Networks}
GNNs~\cite{GNN} have emerged as a powerful class of deep learning models specifically designed to process graph-structured data~\cite{graphdata}. The foundational principle underlying most GNNs is the message-passing mechanism, an iterative process where each node updates its representation by aggregating features from its neighboring nodes. This allows GNNs to capture complex relational patterns and dependencies inherent in graph data, making them highly suitable for applications with linked relationships, such as recommendation systems~\cite{rec2, rec3, rec4}, biomedicine ~\cite{bio2, bio3, bio4}, and social networks~\cite{social2, social3, social4}. Among the influential GNN variants are Graph Convolutional Network (GCN)~\cite{GCN}, which performs normalized aggregation of neighbor features; Graph Attention Network (GAT)~\cite{GAT}, which uses attention mechanisms to weight neighbor contributions dynamically; and Graph Isomorphism Network (GIN)~\cite{GIN}, which offers high expressive power for distinguishing graph structures. Despite their successes, standard GNNs typically require access to the raw graph data for training, which poses significant privacy risks.

To address these risks, various methods~\cite{solitude, GraphPub, blink} have been developed to train GNNs under privacy constraints. These methods can be broadly categorized into those that provide formal privacy guarantees and those that employ heuristic or non-formal privacy techniques. A notable line of work integrates differential privacy directly into the GNN training pipeline. For example, Solitude~\cite{solitude} implements a local differential privacy framework where clients perturb their local adjacency lists and feature vectors using randomized response before sharing them with a server; the server then applies sparsity-based denoising. While providing strong privacy guarantees, its reliance on sparsity assumptions may introduce structural biases. GraphPub~\cite{GraphPub} utilizes a reverse learning process with projected gradient descent and encoder-decoder mechanisms to perturb edges while aiming to maintain graph topology, though its effectiveness is sensitive to parameter selection. Among methods with formal guarantees, Blink~\cite{blink} represents a state-of-the-art approach that uses Bayesian estimation with separated privacy budgets for adjacency lists and degree sequences, achieving high usability but requiring additional privacy budget for degree information.

Conversely, some methods prioritize utility but lack formal privacy proofs. Pri-PGD~\cite{pripgd}, for instance, employs graph forgery through transformations of the graph Laplacian matrix to preserve useful structural information like the convolutional basis spaces and first-order neighborhoods while adding controlled noise. It maintains high model utility with minimal accuracy loss but does not provide a formal differential privacy guarantee.

As summarized in Table~\ref{comp}, existing methods face significant challenges in achieving a balanced performance across all critical criteria, including formal privacy strength, utility preservation on downstream tasks, balance between privacy and utility, and adaptability to graph sparsity. No existing work demonstrates satisfactory performance across all these dimensions simultaneously, highlighting the need for approaches that can better reconcile the inherent tensions between privacy protection and model utility in graph learning.
\section{Preliminaries}

\subsection{Edge Level Differential Privacy}

Edge-level differential privacy~\cite{edgedp} provides a formal mathematical framework for protecting the existence of individual edges in graph-structured data. Two graphs $G$ and $G'$ are considered adjacent if they differ in exactly one edge presence. A randomized mechanism $\mathcal{M}$ satisfies $\epsilon$-edge differential privacy if for all adjacent graphs $G$ and $G'$, and for any possible output set $S \subseteq \text{Range}(\mathcal{M})$, the following inequality holds:

\begin{equation}
\Pr[\mathcal{M}(G) \in S] \leq e^{\epsilon} \cdot \Pr[\mathcal{M}(G') \in S]
\label{eq:edge_dp}
\end{equation}

This definition ensures that the presence or absence of any single edge has limited impact on the output distribution of the mechanism, thereby protecting edge existence from inference attacks. The privacy parameter $\epsilon$ controls the level of protection, with smaller values providing stronger privacy guarantees at the potential cost of reduced data utility.

Differential privacy mechanism ensures that an adversary cannot confidently determine whether any specific edge exists in the original graph based on its output. This protection extends to all edges simultaneously, providing comprehensive privacy coverage for the entire graph structure. The parameter $\epsilon$ quantitatively bounds the privacy loss, as defined by:

\begin{equation}
\mathcal{L} = \max_{S} \left| \ln \frac{\Pr[\mathcal{M}(G) \in S]}{\Pr[\mathcal{M}(G') \in S]} \right| \leq \epsilon
\label{eq:privacy_loss}
\end{equation}
where the maximum is taken over all subsets $S \subseteq \text{Range}(\mathcal{M})$.

This formal guarantee ensures that the maximum information leakage about any individual edge remains controlled within the specified privacy budget, enabling rigorous privacy protection while maintaining the feasibility of meaningful graph analysis tasks.
\subsection{Jaccard Similarity Coefficient}

The Jaccard similarity coefficient~\cite{jaccard}, originally introduced by Paul Jaccard, is a fundamental statistical measure for comparing the similarity and diversity of sample sets. In its most general form, for two finite sets $A$ and $B$, the Jaccard coefficient is defined as the size of their intersection divided by the size of their union:

\begin{equation}
J(A,B) = \frac{|A \cap B|}{|A \cup B|}
\label{eq:jaccard_general}
\end{equation}

This measure produces values in the range $[0,1]$, where $J(A,B) = 0$ indicates completely disjoint sets and $J(A,B) = 1$ signifies identical sets. 

In graph data analysis, the Jaccard coefficient is adapted to measure structural similarity between nodes by considering their neighborhood sets. For any two nodes $i$ and $j$ in a graph $G = (V,E)$, the Jaccard coefficient is defined as:

\begin{equation}
J(i,j) = \frac{|N(i) \cap N(j)|}{|N(i) \cup N(j)|}
\label{eq:jaccard_graph}
\end{equation}

where $N(i)$ and $N(j)$ represent the neighborhood sets of nodes $i$ and $j$ respectively. This adaptation leverages the original set-theoretic foundation of the Jaccard coefficient to capture meaningful topological relationships in graph structures, making it particularly valuable for tasks requiring structural similarity assessment.

It is worth mentioning that the Jaccard coefficient has been used in previous work~\cite{jaccard-p} for link prediction, demonstrating its precise capture of graph structure information. However, that work does not consider GNN training or noisy scenarios, which is different from the setting of our scheme.

\section{Problem Statement}
\label{pb st}
Our goal is to prevent the disclosure of any individual edge in a client's graph while maintaining high node classification accuracy when utilizing a server's resources for GNN training. However, directly sharing the graph adjacency matrix $A$ exposes sensitive edge information to inference attacks by the server. Achieving this privacy-utility balance poses a significant challenge.

A straightforward solution is for the client to perturb the adjacency matrix $A$ locally using randomized response, generating a noisy version $\tilde{A}$ that satisfies $\epsilon$-edge differential privacy before sharing it. While this formal privacy guarantee prevents the server from inferring the existence of any specific edge, the introduced noise often severely distorts the graph's topological structure. Since GNNs heavily rely on the graph structure for message passing and feature aggregation, training on such a perturbed graph $\tilde{A}$ typically leads to significant degradation in node classification accuracy. We must rigorously minimize this performance degradation.

\begin{figure}[htbp]
\centering
\includegraphics[scale=0.56]{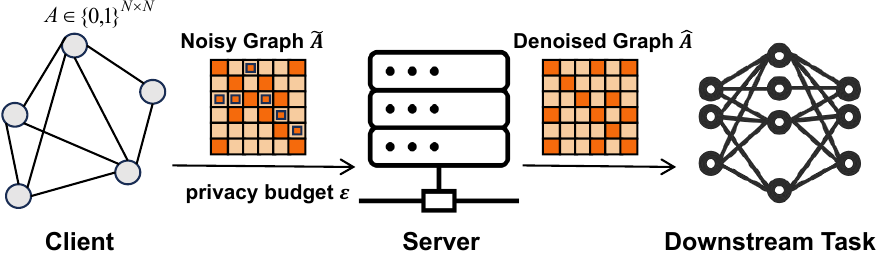}
\caption{System Model}
\label{sys}
\end{figure}

Therefore, the critical problem is to develop a framework that allows the client to share a privacy-preserving graph $\tilde{A}$, while enabling the server to reconstruct a denoised adjacency matrix $\hat{A}$, as shown in Figure~\ref{sys}. This reconstructed graph $\hat{A}$ must preserve the essential structural characteristics necessary for effective GNN training, ensuring high performance in downstream tasks without violating the end-to-end $\epsilon$-edge differential privacy guarantee. 

\section{EdgeRefine}
To achieve an optimal balance between privacy and utility in edge-level differentially private graph training, we propose EdgeRefine . The core innovation involves the server estimating edge existence probabilities in the true graph by computing Jaccard similarity coefficients from the noisy data, followed by an intelligent sampling strategy that selectively retains trustworthy edges. %Specifically, our framework transforms the Jaccard coefficient matrix into connection probabilities through histogram binning, calculates optimal real-to-fake edge ratios based on privacy constraints, and performs stratified sampling according to predetermined sampling rates. 
The subsequent sections provide comprehensive technical details of EdgeRefine. We have also provided additional theoretical analysis of EdgeRefine in Appendix~\ref{the}.

\begin{figure*}[t]
    \centering
    \includegraphics[scale=1]{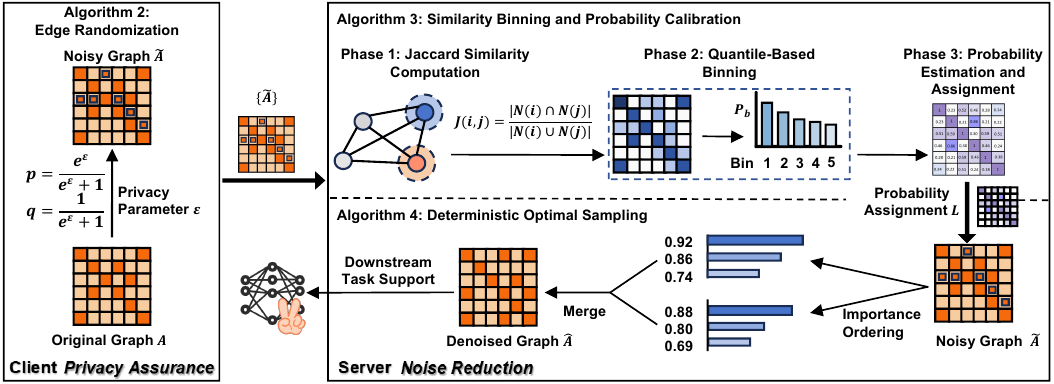}
    \caption{Overall Process of EdgeRefine}
    \label{process}

\end{figure*}

\subsection{Client-Server Pipeline}
\label{subsec:pipeline}
EdgeRefine employs a client-server architecture to execute $\epsilon$-edge differential privacy. The overall process of EdgeRefine is shown in Figure~\ref{process}. In this framework, the client, as the sole owner of the sensitive graph with adjacency matrix $A$ and features $X$, perturbs $A$ locally to produce $\tilde{A}$ satisfying $\epsilon$-edge DP. The server then refines $\tilde{A}$ into a denoised graph $\hat{A}$ suitable for downstream tasks.

Specifically, the end-to-end workflow is formalized in Algorithm~\ref{alg:full_pipeline}.
The client computes the flipping probabilities $p$ and $q$ based on the privacy budget $\epsilon$ to perform randomized response. It then applies this mechanism independently to each undirected edge (by processing the upper triangle of $A$ and mirroring the result) to generate a symmetric perturbed adjacency matrix $\tilde{A}$, which is subsequently transmitted to the server.
The server’s denoising consists of two stages. First, it estimates edge existence probabilities $L$ by computing Jaccard similarities from $\tilde{A}$ and calibrating them via histogram binning. Second, it performs deterministic optimal sampling. Given a sampling rate $k$, the target edge count is $K = \lfloor k \cdot E \rfloor$ (where $E$ is the original edge count). Using $\rho = e^{\epsilon}/(1+e^{\epsilon})$, it selects the top $K_{\text{real}} = \lfloor \rho K \rfloor$ edges from the real candidate set $\mathcal{E}_{\text{real}} = \{(i,j): \tilde{A}_{ij}=1\}$ (edges present after perturbation) and the top $K_{\text{fake}} = K - K_{\text{real}}$ edges from the fake candidate set $\mathcal{E}_{\text{fake}} = \{(i,j): \tilde{A}_{ij}=0\}$ (non-edges or removed edges), with both sets sorted by the estimated probability $L_{ij}$. The union of selected edges forms the final, symmetric matrix $\hat{A}$.

\begin{algorithm}[t]
\caption{Client-Server Pipeline}
\label{alg:full_pipeline}
\begin{algorithmic}[1]
\Require $A \in \{0,1\}^{n \times n}$, $\epsilon > 0$, $k$
\Ensure $\hat{A}$
\Statex
\State \textbf{Client Side:}
\State Generate $\tilde{A} \gets \text{EdgeRandomization}(A, \epsilon)$ \Comment{Algorithm~\ref{alg:rr_mechanism}}
\State Send $\tilde{A}$ to server.
\Statex
\State \textbf{Server Side:}
\State Compute Jaccard matrix $J$ from $\tilde{A}$.
\State Calibrate $J$ to get probability matrix $L$ \Comment{Algorithm~\ref{alg:probability_matrix}}
\State $K \gets \lfloor k \cdot (\frac{1}{2}\sum A_{ij}) \rfloor$, $\rho \gets e^{\epsilon}/(1+e^{\epsilon})$
\State Partition candidates: $\mathcal{E}_{\text{real}} = \{(i,j): \tilde{A}_{ij}=1\}$, $\mathcal{E}_{\text{fake}} = \{(i,j): \tilde{A}_{ij}=0\}$\Comment{ Algorithm~\ref{alg:sampling}}
\State Sort $\mathcal{E}_{\text{real}}$, $\mathcal{E}_{\text{fake}}$ by $L_{ij}$ descending.
\State Select top $\lfloor \rho K \rfloor$ from $\mathcal{E}_{\text{real}}$ and top $K - \lfloor \rho K \rfloor$ from $\mathcal{E}_{\text{fake}}$.
\State Build symmetric matrix $\hat{A}$ from selected edges.
\State \Return $\hat{A}$
\end{algorithmic}
\end{algorithm}

\subsection{Client-side Privacy Assurance}\label{sec5.2}

The client-side perturbation process enforces $\epsilon$-edge differential privacy on the original symmetric adjacency matrix $A \in \{0,1\}^{N \times N}$. The client computes the flipping probabilities for the randomized response mechanism as $p = e^{\epsilon}/(1+e^{\epsilon})$ and $q = 1/(1+e^{\epsilon})$, where $p$ is the probability of retaining a true edge ($A_{ij}=1$) and $q$ is the probability of flipping a non-edge ($A_{ij}=0$) into a fake edge.

To characterize the expected fidelity of the perturbed graph, we define the expected ratio $d$ of real to fake edges in the output $\tilde{A}$. Let $\Lambda = \frac{1}{2}\sum_{i,j} A_{ij}$ denote the count of true edges in the original graph, and $M = N(N-1)/2 - \Lambda$ denote the number of possible non-edges. The expected number of preserved real edges is $p\Lambda$, and the expected number of introduced fake edges is $qM$. Thus, the expected ratio is:
\begin{equation}
d = \frac{p\Lambda}{qM}.
\end{equation}
This ratio $d$ is a function of the privacy budget $\epsilon$ and quantifies the inherent tension between privacy and structural fidelity.

The perturbation is executed as detailed in Algorithm~\ref{alg:rr_mechanism}. The client applies the randomized response once per undirected edge by iterating over the upper triangular part of $A$ (lines 4-8). For each node pair $(i, j)$ with $i < j$, the entry $\tilde{A}_{ij}$ is sampled: $\tilde{A}_{ij} \sim \text{Bernoulli}(A_{ij} = 1 \; ? \; p : q)$. Symmetry is enforced by setting $\tilde{A}_{ji} = \tilde{A}_{ij}$ (line 6), ensuring $\tilde{A}$ represents a valid undirected graph. The resulting matrix $\tilde{A}$, which satisfies $\epsilon$-edge DP, is then sent to the server.

\begin{algorithm}[htbp]
\caption{Edge Randomization}
\label{alg:rr_mechanism}
\begin{algorithmic}[1]
\Require $A \in \{0,1\}^{N \times N}$, privacy parameter $\epsilon > 0$
\Ensure $\tilde{A} \in \{0,1\}^{N \times N}$
\State Initialize $\tilde{A} \gets 0^{N \times N}$
\State $p \gets \frac{e^\epsilon}{e^\epsilon + 1}$, $q \gets \frac{1}{e^\epsilon + 1}$
\For{$i = 1$ to $N$}
    \For{$j = i+1$ to $N$}
        \State $\tilde{A}_{ij} \sim \text{Bernoulli}(A_{ij} = 1 ? p : q)$
        \State $\tilde{A}_{ji} \gets \tilde{A}_{ij}$
    \EndFor
\EndFor
\State \Return $\tilde{A}$
\end{algorithmic}
\end{algorithm}

\subsection{Server-side Noise Reduction}
Upon receiving the noisy adjacency matrix from the client, the server performs a two-stage denoising operation
on it, as shown in Figure~\ref{process}. In general, the server estimates the probability based on limited noisy information, and then performs deterministic sampling on the probability based on the client's privacy budget and their own preset sampling rate. 
\subsubsection{Probability Estimation via Jaccard Similarity and Histogram Binning}
For reconstructing the denoised adjacency matrix, the server first utilizes Jaccard similarity coefficients combined with histogram binning for probability estimation. The fundamental advantage of using Jaccard similarity lies in its normalization by the union of neighborhoods, which naturally suppresses the impact of random edge perturbations. Our implementation calculates the Jaccard coefficient for all node pairs to capture topological relationships. The probability estimation process, outlined in Algorithm~\ref{alg:probability_matrix}, is structured into three sequential phases. 

The initial phase focuses on the computation of the Jaccard similarity matrix for every distinct node pair $(i, j)$ where $i < j$. This involves calculating $J(i,j) = |N(i) \cap N(j)| / |N(i) \cup N(j)|$, where $N(i)$ and $N(j)$ denote the neighborhood sets of nodes $i$ and $j$, respectively. The computation is efficiently performed using optimized set operations, systematically iterating through the upper triangle of the adjacency matrix to avoid redundant calculations.

Following the similarity computation, the second phase employs a quantile-based binning strategy on the set of all pairwise Jaccard coefficients, $\mathcal{X} = \{J(i,j)\mid i < j\}$. This strategy partitions the similarity values into $B$ bins by determining the bin edges at $B+1$ quantiles of the empirical distribution of $\mathcal{X}$. Our quantile-based binning strategy automatically provides higher resolution in densely clustered similarity regions, enhancing discriminative power without manual thresholds or distributional assumptions. This non-parametric binning strategy is robust across diverse graph topologies.

\begin{algorithm}
\caption{Similarity Binning and Probability Calibration}
\label{alg:probability_matrix}
\begin{algorithmic}[1]
\Require 
  \State Adjacency matrix $A \in \{0,1\}^{N \times N}$
  \State Number of bins $B$
\Ensure 
  \State Probability matrix $L \in [0,1]^{N \times N}$
  
\State \textbf{Phase 1: Jaccard Similarity Computation}
\For{$i = 1$ \textbf{to} $N$}
  \For{$j = i+1$ \textbf{to} $N$}
    \State Compute neighborhood sets $N(i)$ and $N(j)$
    \State Calculate$J(i,j) = |N(i) \cap N(j)| / |N(i) \cup N(j)|$
  \EndFor
\EndFor

\State \textbf{Phase 2: Quantile-Based Binning}
\State Extract upper triangle elements $\mathcal{X} = \{J(i,j) \mid i < j\}$
\State Compute bin edges at $B+1$ quantiles of $\mathcal{X}$
\State Initialize bin statistics arrays

\State \textbf{Phase 3: Probability Estimation and Assignment}
\For{each vertex pair $(i,j)$ in upper triangle}
  \State Assign to bin based on Jaccard value
  \State Accumulate edge indicator in bin statistics
\EndFor

\For{each bin $b = 1$ \textbf{to} $B$}
  \State Compute $P_b = \text{sum(edges)} / \text{count(pairs)}$
\EndFor

\For{each vertex pair $(i,j)$}
  \State $L_{ij} = P_b$ for corresponding bin $b$
  \State $L_{ji} = L_{ij}$ \Comment{Ensure symmetry}
\EndFor

\State \Return $L$
\end{algorithmic}
\end{algorithm}

The final phase encompasses probability estimation and the assignment of these probabilities to form the symmetric matrix $L$. Within each bin $b$ defined in the previous phase, the probability $P_b$ is estimated non-parametrically as the ratio of actually existing edges ($A_{ij}=1$) to the total number of node pairs assigned to that bin, $P_b = \text{sum(edges)} / \text{count(pairs)}$. This empirical frequency serves as the calibrated probability of edge existence for all pairs within the bin. These probabilities are then systematically assigned to each vertex pair, with explicit enforcement of symmetry via $L_{ji} = L_{ij}$ to ensure consistency for the undirected graph, resulting in the final probability matrix $L$ that is fed into the subsequent sampling step.

\subsubsection{Deterministic Optimal Sampling with Rate Control}

Given the edge-existence probability matrix $L$ from Algorithm~\ref{alg:probability_matrix}, the server further samples it to obtain a deterministic graph structure. Our deterministic sampling algorithm (Algorithm~\ref{alg:sampling}) ensures precise edge selection and reproducible graph sparsity through two sequential steps. 

\begin{algorithm}
\caption{Deterministic Optimal Sampling}
\label{alg:sampling}
\begin{algorithmic}[1]
\Require 
  \State Probability matrix $L \in [0,1]^{N \times N}$
  \State Sampling rate $k$
  \State Original edge count $E$
\Ensure 
  \State Denoised adjacency matrix $\hat{A} \in \{0,1\}^{N \times N}$
  
\State $K \gets \lfloor k \times E \rfloor$ \Comment{Compute target edge count}
\State $\rho \gets e^{\epsilon} / (1 + e^{\epsilon})$ \Comment{Privacy-utility parameter}

\State \textbf{Candidate Partitioning}
\State $\mathcal{E}_{\text{real}} \gets \{(i,j,L_{ij}) \mid A_{ij} = 1, i < j\}$
\State $\mathcal{E}_{\text{fake}} \gets \{(i,j,L_{ij}) \mid A_{ij} = 0, i < j\}$

\State \textbf{Importance Ordering}
\State $\text{Sort}(\mathcal{E}_{\text{real}}, \text{key}=L_{ij}, \text{order}=\text{descending})$
\State $\text{Sort}(\mathcal{E}_{\text{fake}}, \text{key}=L_{ij}, \text{order}=\text{descending})$

\State \textbf{Deterministic Selection}
\State $K_{\text{real}} \gets \lfloor \rho K \rfloor$
\State $K_{\text{fake}} \gets K - K_{\text{real}}$

\State $\hat{A} \gets \mathbf{0}_{N \times N}$ \Comment{Initialize output matrix}
\State Select top $K_{\text{real}}$ edges from $\mathcal{E}_{\text{real}}$, store in $\mathcal{S}_{\text{real}}$
\State Select top $K_{\text{fake}}$ edges from $\mathcal{E}_{\text{fake}}$, store in $\mathcal{S}_{\text{fake}}$
\State \textbf{Edge Insertion and Symmetrization}
\For{$(i, j)$ in $\mathcal{S}_{\text{real}} \cup \mathcal{S}_{\text{fake}}$}
    \State $\hat{A}_{ij} \gets 1$, $\hat{A}_{ji} \gets 1$ \Comment{Insert edge and enforce symmetry}
\EndFor

\State \Return $\hat{A}$
\end{algorithmic}
\end{algorithm}
The first stage handles initialization, candidate set partitioning, and sorting. Our algorithm begins by calculating the target number of edges to preserve, $K = \lfloor k \times E \rfloor$, which is controlled by the global sampling rate $k$ to explicitly govern the sparsity of the denoised graph. Simultaneously, the privacy-utility trade-off parameter $\rho = e^{\epsilon}/(1+e^{\epsilon})$ is derived from the privacy budget $\epsilon$. All potential node pairs are then partitioned into two candidate sets based on the observed noisy adjacency matrix $\tilde{A}$: the real candidate set $\mathcal{E}_{\text{real}}$ for observed edges ($\tilde{A}_{ij}=1$) and the fake candidate set $\mathcal{E}_{\text{fake}}$ for non-edges ($\tilde{A}_{ij}=0$), with each pair associated with its estimated existence probability $L_{ij}$ from the probability matrix. Both candidate sets are subsequently sorted in descending order of their $L_{ij}$ values. This deterministic process, termed Importance Ordering, eliminates any sampling variance and ensures reproducible intermediate results.

The second stage performs deterministic selection and merging. The algorithm selects the top $K_{\text{real}} = \lfloor \rho K \rfloor$ edges from the sorted real set $\mathcal{E}_{\text{real}}$ and the top $K_{\text{fake}} = K - K_{\text{real}}$ edges from the sorted fake set $\mathcal{E}_{\text{fake}}$. This provides an optimal balance, governed by $\rho$, between preserving high-confidence observed edges and carefully recovering potentially missing true edges. Finally, our algorithm initializes a zero matrix, populates it with edges from both candidate sets, and symmetrizes it to obtain the denoised adjacency matrix $\hat{A}$. The key advantages of this design are guaranteed reproducibility and the complete elimination of sampling variance.

The denoised adjacency matrix $\hat{A}$ we have obtained so far can effectively support downstream tasks. Theoretical analysis of security and utility for denoised matrix will be presented in Appendix ~\ref{the}, while the experimental results in Section~\ref{ex} and Appendix~\ref{ex-exp}.

\section{Experimental Evaluation}
\label{ex}
In this section, we first introduce the experimental settings. Then, we answer the following research questions through extensive experiments to demonstrate the superior performance of EdgeRefine.

\begin{itemize}
\item \textbf{RQ1: } Does EdgeRefine outperform existing privacy preserving graph learning methods in terms of privacy-utility balance across different GNN architectures (GAT, GCN, GIN) and diverse graph datasets?
\item \textbf{RQ2: } Does EdgeRefine have a significant improvement in sparsity compared to other baselines, and does this improvement in sparsity bring computational advantages?
\item \textbf{RQ3: }  What is the level of error in probability estimation, and does the overall process depend on the accuracy of probability estimation?
\item \textbf{RQ4: } What is the impact of key parameters, particularly the sampling rate, on EdgeRefine's performance?
\item \textbf{RQ5: } How do the individual components of EdgeRefine contribute to its overall effectiveness (ablation study)?
\item \textbf{RQ6: } Can EdgeRefine still maintain low performance degradation relative to the noise-free baseline in unconventional differential privacy graph learning scenarios?
\item \textbf{RQ7: }When faced with a state-of-the-art graph-reconstruction attack, can EdgeRefine still prevent edge information from being leaked?
\end{itemize}

\subsection{Experimental Settings  }
\label{experisetting}

\noindent\textbf{Hardware Settings: } The evaluation is performed on a desktop with 11th Gen Intel(R) Core(TM) i7-11700 @2.50GHz processor and 32GB of running memory. 

\noindent\textbf{Software Environment: } The software ecosystem was built upon the Python programming language. Key libraries utilized for the experiment included NumPy~\cite{numpy} for foundational numerical computations and multi-dimensional array operations, PyTorch~\cite{pytorch} for deep learning model construction and training, and Scikit-learn~\cite{scikit-learn} for implementing traditional machine learning algorithms and data preprocessing. SciPy~\cite{SciPy} was employed for scientific computing and advanced statistical functions. Furthermore, specialized libraries such as torch\_geometric~\cite{pyg} were used for handling graph-structured data, and networkx~\cite{networkx} was used for network analysis and manipulation. Unless otherwise specified, the main node-classification accuracy results are averaged over five independent runs, while all other experiments are averaged over three independent runs.

\noindent\textbf{GNN Architecture: } We used three different GNN architectures (GAT, GCN, and GIN) in the experiments to demonstrate that EdgeRefine is applicable to different message passing processes. We have included the specific process of aggregating neighbor information using these architectures in Appendix~\ref{gn}.

\noindent\textbf{Datasets: } We evaluated EdgeRefine on four real-world graph datasets for node classification and one real-world dataset for graph classification. These datasets all originate from real-world association networks: citation networks (e.g., Cora~\cite{cora}) reflect academic inheritance via paper citations; product networks (e.g., AMAP~\cite{AMAP}) capture user-item links through co-purchases; and author/paper collaborations (e.g., ACM~\cite{acm}, DBLP~\cite{dblp}) model research partnerships via co-authorship, with features including content-based representations by binary word vectors. 
MUTAG~\cite{mutag} is a dataset containing 188 chemical molecule graphs, with an average of 17.9 nodes and 7 atomic type labels per graph, commonly used for graph classification tasks. Due to the essential differences in data structure and task objectives between multi graph datasets and single graph datasets, Table~\ref{tab:datasets_nice} only summarizes the statistical characteristics of four datasets for node classification.
\begin{table}[htbp]
\centering
\caption{ Statistics of Graph Datasets}
\label{tab:datasets_nice}
\begin{tabular*}{\columnwidth}{@{\hspace{6pt}} l@{\extracolsep{\fill}} ccccc @{\hspace{2pt}}}
\toprule
\textbf{Dataset} & \textbf{Samples} & \textbf{Dimension} & \textbf{Edges} & \textbf{Classes} \\
\midrule
{ACM\cite{acm}} & 3025 & 1870 & 13128 & 3 \\
{DBLP\cite{dblp}} & 4057 & 334 & 3528 & 4 \\
{AMAP\cite{AMAP}} & 7650 & 745 & 119081 & 8 \\
{Cora\cite{cora}} & 2708 & 1433 & 5278 & 7 \\
\bottomrule
\end{tabular*}
\end{table}

\noindent\textbf{Baselines: }To demonstrate the performance of EdgeRefine, we implemented the following state-of-the-art and high-performance solutions as baselines for comparison:

\begin{itemize}

    \item \textbf{Blink~\cite{blink}}: This framework includes three variants: hard, soft, and hybrid. Blink-hard variant applies a threshold to the Bayesian posterior probability matrix, preserving only edges exceeding a predefined cutoff, providing effective noise removal under low privacy budgets but potentially discarding genuine edges. The soft approach uses edge probabilities as weights without thresholding, preserving more information but being susceptible to noise. Blink-hybrid strategy combines thresholding and weighting, first filtering edges then using their probabilities as weights, offering flexibility but requiring careful parameter tuning. Based on the original data, both hard and hybrid variants demonstrate superior performance under low privacy budgets, making them the preferred choices for comparisons in our experiments. 
    
    \item \textbf{DPRR~\cite{dprr}}: It employs Warner's Randomized Response on each edge, followed by an edge sampling process with probabilities calibrated using node degrees perturbed by the Laplace mechanism. This method focuses on preserving the degree distribution of the graph after perturbation under edge LDP guarantees.
    
    \item \textbf{LDPGen~\cite{ldpgen}}: Operating under the local DP model, users perturb and upload their degree vectors via the Laplace mechanism. The server then clusters users based on these noisy vectors and generates a synthetic graph using a generative model like BTER, rather than releasing a directly perturbed adjacency matrix.
    
    \item \textbf{LAPGRAPH~\cite{lapgraph}}: This approach adds Laplace noise directly to each element of the adjacency matrix. It then constructs a private, sparse adjacency matrix by selecting the top-T edges with the largest noisy values. The method aims to preserve graph sparsity but may introduce significant noise in large graphs.
\end{itemize}

To establish performance baselines, we also compared EdgeRefine against noise-free GNN implementations marked as \textbf{Origin}, which provides theoretical upper bounds for utility preservation. We also evaluated against standard differentially private edge sampling methods to demonstrate the advantages of our denoising approach over simple perturbation techniques.

\noindent\textbf{Data Preparation and Model Setting: }
To comprehensively evaluate the performance of EdgeRefine, we conducted extensive experiments under rigorous privacy constraints. 

\textit{Data Preparation.} For all datasets, we randomly split the nodes into training, validation, and test sets with a ratio of 2:1:1, ensuring balanced distribution across all subsets. This partitioning strategy maintains sufficient training data while providing adequate samples for validation and reliable performance estimation.

\textit{Privacy Setting.} The privacy protection mechanism follows a two-stage process: client-side perturbation using randomized response with privacy budget $\epsilon$, followed by server-side denoising through Jaccard-based probability estimation and deterministic sampling. We evaluated our approach across a range of privacy budgets $\epsilon \in \{0.5, 1.0, 1.5, 2.0, 2.5, 3.0, 3.5\}$ to systematically analyze the privacy-utility trade-off.

\textit{Model Selection. } To ensure robust performance assessment, we employed multiple state-of-the-art graph neural network architectures including GCN~\cite{GCN},GAT~\cite{GAT}, and GIN~\cite{GIN}. This architectural diversity allows us to evaluate the generalizability of our privacy mechanism across different message-passing paradigms. For each combination of dataset, GNN architecture, and privacy budget, we conducted grid search to identify optimal hyperparameters based on validation performance.

For all methods, we used the Adam optimizer with a learning rate of 0.005 and a weight decay of 0.00001. Early stopping was applied with a patience of 100 epochs based on validation accuracy. All experiments were run for at most 700 epochs, with performance evaluated every 50 epochs. For EdgeRefine, the sampling rate was fixed at $k=0.01$ unless otherwise specified. For Blink, the privacy budget allocated to the degree sequence was fixed at $\epsilon_d=0.1$.

\textbf{Evaluation Metrics:} 
In downstream tasks, model performance serves as a key indicator of the usability of differentially private data. To evaluate the privacy-utility trade-off of EdgeRefine, we adopt node classification as the benchmark task, measuring model performance across varying privacy budgets with classification accuracy as the primary metric. We assess performance on the test set, reporting mean values over five independent runs to ensure statistical significance. Since the selected privacy budget range ($\epsilon \in [0.5, 3.5]$) is relatively low, achieving competitive classification accuracy within this range indicates a favorable privacy-utility balance.

Beyond accuracy, we also introduce the following metrics to evaluate the performance of EdgeRefine, including stability, attack resistance, and so on.

\textit{(1) Stability Assessment:}
Two statistical measures: variance ($\sigma^2$) and coefficient of variation ($CV$) are introduced to evaluate the stability and consistency of different methods under varying privacy constraints. Variance quantifies the dispersion of accuracy values across privacy budgets, with lower values indicating more stable performance. The coefficient of variation, calculated as $CV = \sigma/\mu$ (where $\sigma$ is the standard deviation and $\mu$ is the mean accuracy), provides a normalized measure of variability, facilitating comparison across methods with different accuracy scales. These stability metrics are crucial for assessing the practical reliability of privacy-preserving methods in real-world scenarios where privacy budgets may fluctuate. In Appendix~\ref{ex-avg}, we also use std to evaluate stability.

\textit{(2) Privacy-Utility Balance Index (PUBI):}
Existing evaluations typically focus on absolute accuracy at isolated privacy budgets, which overlooks a critical dimension: whether a method maintains stable utility as privacy requirements are relaxed. A method that excels under stringent privacy (low $\epsilon$) but deteriorates when the budget increases, it fails to achieve a genuine privacy-utility balance. To quantify this property, we define the Privacy-Utility Balance Index (PUBI). Let $\mathrm{Acc}(\epsilon_i)$ denote the node classification accuracy of EdgeRefine at privacy budget $\epsilon_i$ and $\mathrm{Acc_{ori}}$ the accuracy of the noise-free baseline. We first define the Average Utility Retention (AUR):
\begin{equation}
\mathrm{AUR} = \frac{1}{N}\sum_{i=1}^{N} \frac{\mathrm{Acc}(\epsilon_i)}{\mathrm{Acc_{ori}}},
\label{eq-aur}
\end{equation}
which measures the mean utility preserved across the privacy budget spectrum. 

Second, we define the Monotonic Consistency Factor (MCF):

\begin{equation}
\mathrm{MCF} = 1 - \frac{1}{N-1}\sum_{i=1}^{N-1} \max\!\left(0, \frac{\mathrm{Acc}(\epsilon_i) - \mathrm{Acc}(\epsilon_{i+1})}{\mathrm{Acc_{ori}}}\right), 
\end{equation}
which penalizes performance degradation when the privacy budget increases (i.e., privacy protection weakens). The final index of PUBI is their product:
\begin{equation}
\mathrm{PUBI} = \mathrm{AUR} \times \mathrm{MCF}.
\label{eq-pubi}
\end{equation}
PUBI rewards both high overall utility and stable behavior across privacy regimes. 
A method that suffers systematic decay as $\epsilon$ increases receives a lower score despite strong performance at isolated low budgets.

\textit{(3) Probability Estimation Quality:}
We also evaluate the probability estimation error using three indicators  to demonstrate the effectiveness of probability estimation.  The smaller the three error indicators, the more accurate the estimation. 
We introduce Mean Absolute Error (MAE), Brier score,  Expected Calibration Error(ECE)/probability Mean Absolute Deviation (pMAD). The MAE measures the average absolute difference between estimated probabilities and actual binary edge indicators. The Brier score quantifies the mean squared error of probability predictions. The ECE assesses the calibration of probability estimates by comparing expected accuracy with predicted confidence. The pMAD represents the dispersion of probability estimates around their mean. The ECE/pMAD ratio indicates how calibration errors relate to inherent variability; a lower ratio implies calibration errors are small relative to the natural spread of estimates.

\textit{(4) Resistance to Reconstruction Attacks:}
To evaluate privacy protection against inference attacks, we employ the Relative Absolute Error (RAE) metric. RAE quantifies an attacker's ability to recover original edges from the perturbed graph, with higher RAE values indicating greater reconstruction error and thus stronger resistance. RAE is defined as:
\begin{equation}
\text{RAE} = \frac{\sum_{i,j} |A_{ij} - \hat{A}_{ij}|}{\sum_{i,j} |A_{ij} - \bar{A}|},
\end{equation}
where $A_{ij}$ is the true adjacency matrix, $\hat{A}_{ij}$ is the reconstructed matrix, and $\bar{A}$ is the mean of the true adjacency matrix entries. An RAE close to 1 indicates the attack performs no better than guessing the mean, while an RAE > 1 suggests the attack performs worse, demonstrating strong resilience. Thus, higher RAE indicates better privacy protection.

\subsection{Experimental Results and Analysis}

\begin{figure*}[t]
    \centering
    \includegraphics[width=\textwidth]{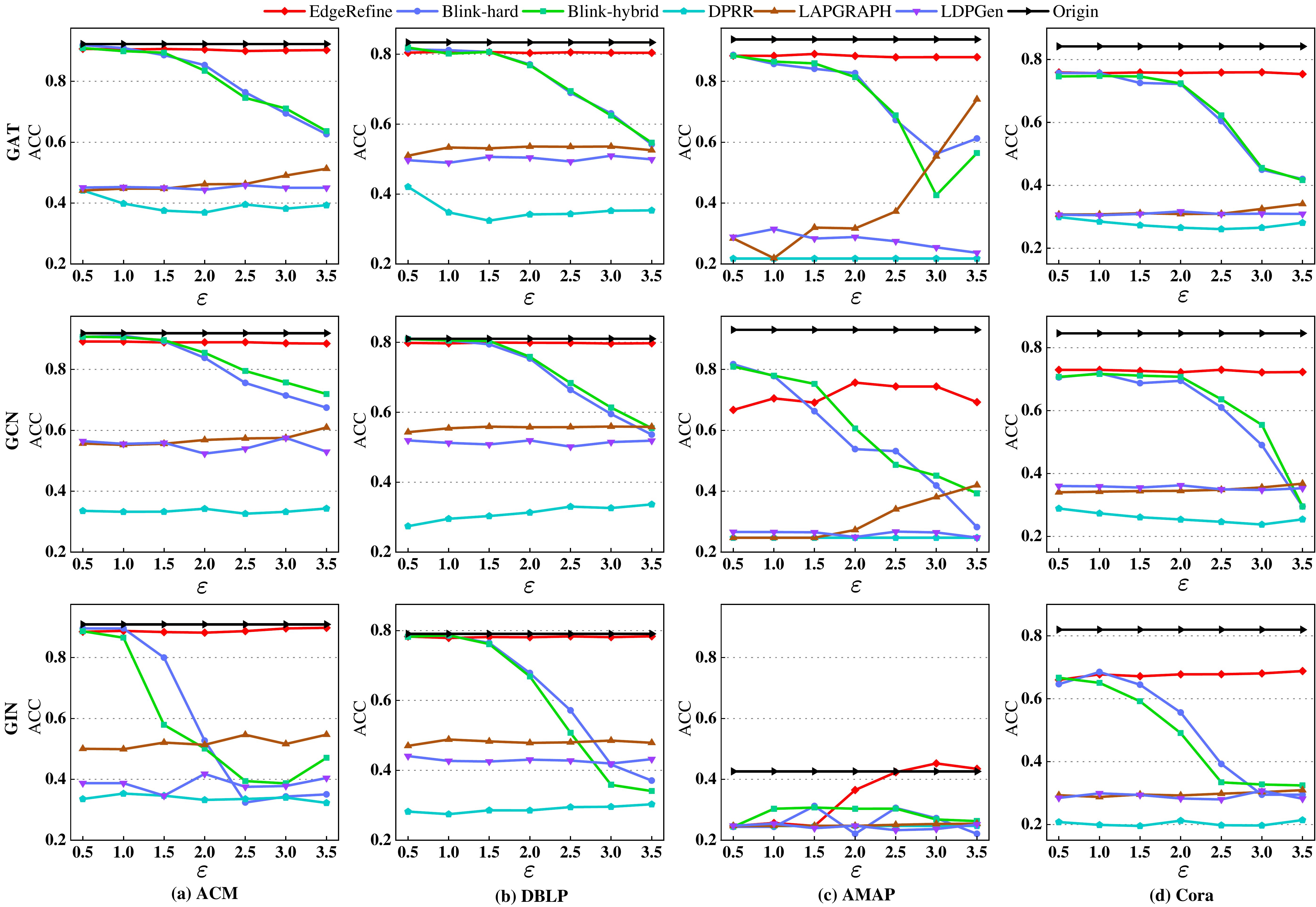}
    \caption{Performance Comparison of EdgeRefine and Baseline Methods}
    \label{exp}
\end{figure*}

\subsubsection{Privacy-utility of EdgeRefine (\textbf{RQ1})}
\label{total exp}
To evaluate privacy-utility trade-off of EdgeRefine and current state-of-the-art baselines, we first conducted experiments comparing their performance across different privacy budgets and four distinct datasets. We present the node classification accuracy in Figure~\ref{exp}, the variance and coefficient of variation of mode accuracy in Table~\ref{tab:cv_variance}, and privacy-utility balance index in Table~\ref{tab:pubi}.

On the ACM and DBLP datasets shown in Figures~\ref{exp}(a) and ~\ref{exp}(b), EdgeRefine exhibits exceptional consistency, maintaining accuracy close to noise-free baseline.  On the ACM dataset, it achieves accuracy of around 0.90–0.91 for GAT, 0.88–0.89 for GCN, and 0.88–0.90 for GIN throughout the privacy budget spectrum. On the DBLP dataset, GAT, GCN and GIN also maintain stable accuracy ranging from 0.77 to 0.81. EdgeRefine is generally superior to existing baselines on both datasets under low privacy budgets and is close to noise-free baseline. It is worth noting that at the lowest privacy budget ($\epsilon=0.5$) on the ACM dataset, Blink-hard achieves slightly higher accuracy (around 0.92 for GAT) compared to EdgeRefine (around 0.91 for GAT at $\epsilon=0.5$). However, Blink-hard shows significant performance degradation with increasing $\epsilon$. When the privacy budget is 2.5, EdgeRefine has improved GAT performance by 17.8\% compared to Blink on the ACM dataset. The performance degradation in Blink methods is attributed to higher privacy budgets introducing lower noise levels, which causes more edges to be estimated as high-probability edges, consequently leading to the selection of more edges, increased graph density, and ultimately impaired performance. Other baseline methods show consistently low accuracy across all experimental conditions.

The AMAP dataset evaluation shows that EdgeRefine maintains strong performance with GAT accuracy ranging from 0.88--0.89 and GCN accuracy from 0.67--0.76 across privacy budgets in Figure~\ref{exp}(c). Notably, the GIN architecture exhibits significantly lower performance around 0.25--0.45. Without privacy protection, the GIN baseline accuracy is already low (0.4261), and at higher privacy budgets, the performance abnormally exceeds the Origin baseline. 
The limited performance of the GIN model on the AMAP dataset reflects an inherent instability of this dataset, corroborated by Appendix~\ref{ex-avg}, where elevated standard deviations appear across all three architectures. Our controlled experiment in Appendix~\ref{ex-uno} further confirms that EdgeRefine at $\epsilon=3.0$ does not differ significantly from the noise-free baseline ($p = 0.226$); the apparent performance gains at higher privacy budgets are therefore attributable to sampling fluctuations within GIN's broad variance band, not to a reliable regularization effect of DP noise. Overall, EdgeRefine provides more stable performance across different privacy budgets and networks, whereas other baseline methods demonstrate inferior performance.

On the Cora dataset, as shown in Figure~\ref{exp}(d) EdgeRefine demonstrates consistent performance across privacy budgets, with accuracy maintaining stable levels around 0.75--0.76 for GAT, 0.72--0.73 for GCN, and 0.66--0.69 for GIN as $\epsilon$ increases from 0.5 to 3.5. In contrast, Blink-hard exhibits a clear declining trend, while Blink-hybrid shows similar deterioration. The performance deterioration is particularly evident for GCN architecture. When the privacy budget is 2.5, EdgeRefine improves GCN performance by 19.7\% compared to Blink on the Cora dataset. Other baseline methods maintain consistently low accuracy below 0.40 across all privacy settings.

To show the stability of EdgeRefine, Table~\ref{tab:cv_variance} summarizes the coefficient of variation (CV) and variance ($\sigma^2$) computed from the updated experimental results. EdgeRefine achieves the lowest variance on ACM ($\sigma^2=0.0001$, CV=0.01) and DBLP ($\sigma^2=0.0001$, CV=0.01), where its stability markedly exceeds that of all baselines. On Cora ($\sigma^2=0.0012$, CV=0.05), it remains competitive with the most stable alternatives. AMAP stands out as a distinct case: its structural properties amplify variance across all methods (see Appendix~\ref{ex-avg}), and while DPRR and LDPGen exhibit lower CV, their absolute accuracies remain substantially below EdgeRefine's. Notably, Blink-hard and Blink-hybrid show the highest variance on every dataset. Additionally, the standard deviation from five independent runs per configuration is provided in Appendix~\ref{ex-avg}. These results confirm that EdgeRefine maintains superior stability with minimal performance fluctuation on the majority of datasets.

\begin{table}[htbp]
\centering
\caption{Coefficient of Variation (CV) and Variance  ($\sigma^2$) }
\label{tab:cv_variance}

 \small
\setlength{\tabcolsep}{3pt}

\begin{tabular*}{\columnwidth}{@{\hspace{4pt}} p{0.2\columnwidth} @{\extracolsep{\fill}}  *{4}{c} @{\hspace{4pt}}}

\toprule
\multirow{2}{*}{\textbf{Scheme}} & \textbf{ACM} & \textbf{DBLP} & \textbf{AMAP} & \textbf{Cora} \\
 & CV/$\sigma^2$ & CV/$\sigma^2$ & CV/$\sigma^2$ & CV/$\sigma^2$ \\
\midrule
EdgeRefine & 0.01/0.0001 & 0.01/0.0001 & 0.36/0.0530 & 0.05/0.0012 \\
Blink-hard & 0.26/0.0374 & 0.19/0.0174 & 0.45/0.0572 & 0.27/0.0245 \\
Blink-hybrid & 0.23/0.0301 & 0.21/0.0205 & 0.42/0.0528 & 0.27/0.0244 \\
DPRR & 0.09/0.0009 & 0.11/0.0012 & 0.06/0.0002 & 0.14/0.0011 \\
LAPGRAPH & 0.09/0.0022 & 0.06/0.0010 & 0.38/0.0149 & 0.07/0.0005 \\
LDPGen & 0.15/0.0048 & 0.08/0.0014 & 0.08/0.0004 & 0.09/0.0008 \\
\bottomrule
\end{tabular*}
\begin{tablenotes}
\small
\item \textbf{Notes:} The final result of each method is calculated by taking the average of GAT, GCN, and GIN.
\end{tablenotes}
\end{table}

Furthermore, Table~\ref{tab:pubi} reports PUBI for all six methods across the 12 architecture--dataset combinations. EdgeRefine consistently achieves the highest PUBI across all combinations, with an average of 0.9386. By contrast, Blink-hybrid and Blink-hard average only 0.7101 and 0.6902, respectively, trailing EdgeRefine by 24.3\% and 26.4\%. The divergence stems from MCF. Blink-hard and Blink-hybrid occasionally outperform EdgeRefine at the strictest privacy levels ($\epsilon=0.5$); however, their accuracy decays sharply as $\epsilon$ grows. For example, on GAT-AMAP, Blink-hard plummets from 0.8856 to 0.2739 across the budget range, yielding MCF$=0.8912$, whereas EdgeRefine remains stable (0.8582--0.8922) with MCF$=0.9955$. This systemic fragility undermines their overall balance. 
Notably, on GIN-AMAP, EdgeRefine attains PUBI (1.0921)$>1$   because its measured accuracy occasionally exceeds the noise-free baseline. As analyzed in Appendix~\ref{ex-uno}, this apparent improvement is attributable to the high variance of GIN on the AMAP dataset.

\begin{table}[t]
\caption{Privacy-Utility Balance Index Across All Architecture--Dataset Combinations}
\label{tab:pubi}

\centering

\footnotesize
\setlength{\tabcolsep}{1.65pt}

\begin{tabular*}{\columnwidth}{@{\hspace{1pt}} p{0.08\columnwidth} @{\extracolsep{\fill}} l *{6}{c} @{\hspace{1pt}}}

\toprule
Arch. & Dataset & \textbf{EdgeRefine} & Blink-hard & Blink-hybrid & DPRR & LAPGRAPH & LDPGen \\
\midrule
\multirow{4}{*}{GAT}
& ACM  & \textbf{0.9779} & 0.8118 & 0.7556 & 0.4571 & 0.4839 & 0.4868 \\
& AMAP & \textbf{0.9289} & 0.6481 & 0.7674 & 0.2864 & 0.4145 & 0.2594 \\
& Cora & \textbf{0.8958} & 0.6860 & 0.6852 & 0.3348 & 0.3785 & 0.3636 \\
& DBLP & \textbf{0.9711} & 0.8384 & 0.8136 & 0.4183 & 0.6349 & 0.5931 \\
\midrule
\multirow{4}{*}{GCN}
& ACM  & \textbf{0.9788} & 0.8161 & 0.8470 & 0.3826 & 0.5802 & 0.5268 \\
& AMAP & \textbf{0.8362} & 0.5601 & 0.6055 & 0.2778 & 0.3883 & 0.2891 \\
& Cora & \textbf{0.8284} & 0.6560 & 0.6973 & 0.3215 & 0.4127 & 0.4233 \\
& DBLP & \textbf{0.9911} & 0.8120 & 0.8449 & 0.3744 & 0.6756 & 0.6298 \\
\midrule
\multirow{4}{*}{GIN}
& ACM  & \textbf{0.9775} & 0.6258 & 0.6771 & 0.3855 & 0.5206 & 0.4283 \\
& AMAP & \textbf{1.0921} & 0.5953 & 0.5505 & 0.5910 & 0.5973 & 0.5772 \\
& Cora & \textbf{0.7915} & 0.5476 & 0.5790 & 0.2662 & 0.3753 & 0.3569 \\
& DBLP & \textbf{0.9945} & 0.6854 & 0.6984 & 0.3493 & 0.6069 & 0.5262 \\
\midrule
\multicolumn{2}{@{}l}{Average} & \textbf{0.9386} & 0.6902 & 0.7101 & 0.3704 & 0.5057 & 0.4550 \\
\bottomrule

    \end{tabular*}
\end{table}

\subsubsection{Sparsity Analysis (\textbf{RQ2})}
\label{ex-sp}
To evaluate the sparsity maintenance performance of our framework, we used GAT network to test the graph density and training time of all methods on different privacy budgets on the DBLP dataset. Specifically, Blink-hybrid baseline does not appear in this experiment due to the inability to uniformly test using edge weights as adjacency matrices. Our test results are presented in Tables ~\ref{tab:density_values_transposed} and ~\ref{tab:training_time_transposed}.

The density analysis in Table~\ref{tab:density_values_transposed} demonstrates that EdgeRefine consistently maintains superior sparsity preservation across all privacy budgets. Notably, EdgeRefine achieves density values in the range of \(10^{-6}\) to \(10^{-5}\), representing a 2-5 order of magnitude improvement over alternative approaches. This exceptional sparsity preservation is particularly evident at lower privacy budgets (\(\epsilon \in [0.5, 1.5]\)), where EdgeRefine outperforms all comparative methods. The Blink-hard method shows competitive performance at minimal privacy budgets but exhibits significant density inflation as \(\epsilon\) increases beyond 2.0, reaching density values approximately 1000× higher than EdgeRefine at \(\epsilon = 3.5\). In contrast, methods like LDPGen and DPRR maintain substantially denser graphs across all privacy settings, with density values consistently above \(10^{-2}\), indicating limited applicability in sparsity-sensitive applications. We also analyzed the relationship between density and accuracy in this dataset, as detailed in Appendix~\ref{ex-de}.

\begin{table}[htbp]
\centering
\caption{Normalized Density Values Across Methods}
\label{tab:density_values_transposed}

\footnotesize
\setlength{\tabcolsep}{3pt}

\begin{tabular*}{\columnwidth}{@{\hspace{3pt}} p{0.075\columnwidth} @{\extracolsep{\fill}} l *{5}{c} @{\hspace{1pt}}}

\toprule
\textbf{$\epsilon$} & \textbf{EdgeRefine} & \textbf{Blink-hard} & \textbf{LDPGen} & \textbf{DPRR} & \textbf{LAPGRAPH} \\
\midrule
0.5 & \(4.255 \times 10^{-6}\) & \(\mathbf{1.216 \times 10^{-7}}\) & \(6.129 \times 10^{-1}\) & \(3.778 \times 10^{-1}\) & \(3.777 \times 10^{-1}\) \\
1.0 & \(4.255 \times 10^{-6}\) & \(\mathbf{8.510 \times 10^{-7}}\) & \(4.659 \times 10^{-1}\) & \(2.695 \times 10^{-1}\) & \(2.691 \times 10^{-1}\) \\
1.5 & \(\mathbf{4.255 \times 10^{-6}}\) & \(4.923 \times 10^{-5}\) & \(3.321 \times 10^{-1}\) & \(1.827 \times 10^{-1}\) & \(1.829 \times 10^{-1}\) \\
2.0 & \(\mathbf{4.255 \times 10^{-6}}\) & \(5.668 \times 10^{-4}\) & \(2.247 \times 10^{-1}\) & \(1.196 \times 10^{-1}\) & \(1.195 \times 10^{-1}\) \\
2.5 & \(\mathbf{4.255 \times 10^{-6}}\) & \(2.602 \times 10^{-3}\) & \(1.465 \times 10^{-1}\) & \(7.599 \times 10^{-2}\) & \(7.609 \times 10^{-2}\) \\
3.0 & \(\mathbf{4.255 \times 10^{-6}}\) & \(6.457 \times 10^{-3}\) & \(9.337 \times 10^{-2}\) & \(4.784 \times 10^{-2}\) & \(4.794 \times 10^{-2}\) \\
3.5 & \(\mathbf{4.255 \times 10^{-6}}\) & \(1.056 \times 10^{-2}\) & \(5.842 \times 10^{-2}\) & \(2.956 \times 10^{-2}\) & \(2.963 \times 10^{-2}\) \\
\bottomrule
\end{tabular*}
\end{table}

Table~\ref{tab:training_time_transposed} presents the training and preprocessing time for all baselines. EdgeRefine achieves the fastest training time across most privacy budgets, with execution time consistently below 4 milliseconds. This represents a 3-4 order of magnitude improvement over methods like LDPGen, which requires thousands of milliseconds to complete training. The exceptional efficiency of EdgeRefine can be attributed to its ability to maintain ultra-sparse graph structures, which dramatically reduces the computational overhead of neighborhood aggregation operations in GNNs. Notably, Blink-hard demonstrates competitive training time at lower privacy budgets but experiences exponential time complexity growth as \(\epsilon\) increases, with training time rising to 62.3 milliseconds at \(\epsilon = 3.5\), while EdgeRefine maintains a steady training time.

\begin{table}[htbp]
\centering
\caption{Training Time (ms) and Preprocessing Time (s) Across Methods}
\label{tab:training_time_transposed}
\small
\setlength{\tabcolsep}{2.5pt}

\begin{tabular}{@{\hspace{3pt}} p{0.075\columnwidth} @{\extracolsep{\fill}} l *{5}{c} @{\hspace{1pt}}}
\toprule
\textbf{$\epsilon$} & \textbf{EdgeRefine} & \textbf{Blink-hard} & \textbf{LDPGen} & \textbf{DPRR} & \textbf{LAPGRAPH} \\
\midrule
0.5 & \(\mathbf{1.572 }\) (1104.3) & \(1.938 \) (46.5) & \(3644 \) (1.9) & \(790.8 \) (13.5) & \(734.5 \) (1.1) \\
1.0 & \(1.420 \) (836.1) & \(\mathbf{1.256 }\) (46.5) & \(1922 \) (1.9) & \(345.8 \) (12.7) & \(350.7 \) (1.2) \\
1.5 & \(1.891 \) (628.4) & \(\mathbf{1.205 }\) (46.5) & \(425.7 \) (1.9) & \(242.7 \) (12.1) & \(242.2 \) (1.2) \\
2.0 & \(\mathbf{1.274 }\) (491.1) & \(3.513 \) (46.5) & \(293.4 \) (1.9) & \(375.5 \) (11.6) & \(364.6 \) (1.2) \\
2.5 & \(\mathbf{1.348 }\) (427.8) & \(14.15 \) (46.6) & \(189.1 \) (1.9) & \(123.7 \) (11.3) & \(212.6 \) (1.2) \\
3.0 & \(\mathbf{3.410 }\) (357.6) & \(25.35 \) (46.5) & \(253.8 \) (1.9) & \(64.13 \) (11.0) & \(126.4 \) (1.1) \\
3.5 & \(\mathbf{2.554 }\) (308.3) & \(62.30 \) (46.6) & \(176.0 \) (1.9) & \(78.71 \) (10.9) & \(156.3 \) (1.1) \\
\bottomrule

\end{tabular}

\begin{tablenotes}
\small
\item \textbf{Notes:}  Preprocessing time in parentheses 
\end{tablenotes}

\end{table}

Regarding preprocessing time (shown in parentheses in Table~\ref{tab:training_time_transposed}), EdgeRefine exhibits relatively longer preprocessing time, ranging from 308.3 to 1104.3 seconds, which is due to the iterative edge sampling and refinement process. However, it is important to emphasize that the preprocessing module of EdgeRefine is completely decoupled from the downstream training phase. In practical deployment scenarios, preprocessing tasks can be executed on dedicated servers and completed offline in advance. Therefore, they exert no substantial impact on the overall system performance. In contrast, methods like DPRR, LAPGRAPH, and LDPGen show much shorter preprocessing time, but their training time becomes significantly longer, making them less efficient for practical use.

The inverse relationship between privacy protection strength and computational efficiency follows expected theoretical patterns, with stricter privacy guarantees (lower $\epsilon$) generally requiring more intensive processing. EdgeRefine incurs the longest preprocessing time but maintains strong training efficiency across the privacy budget spectrum, effectively decoupling training cost from privacy strength while achieving high performance.

\subsubsection{Probability Estimation Quality Analysis (\textbf{RQ3})}
\label{error ex}
\begin{figure}[htbp]
    \centering
    \includegraphics[width=\columnwidth]{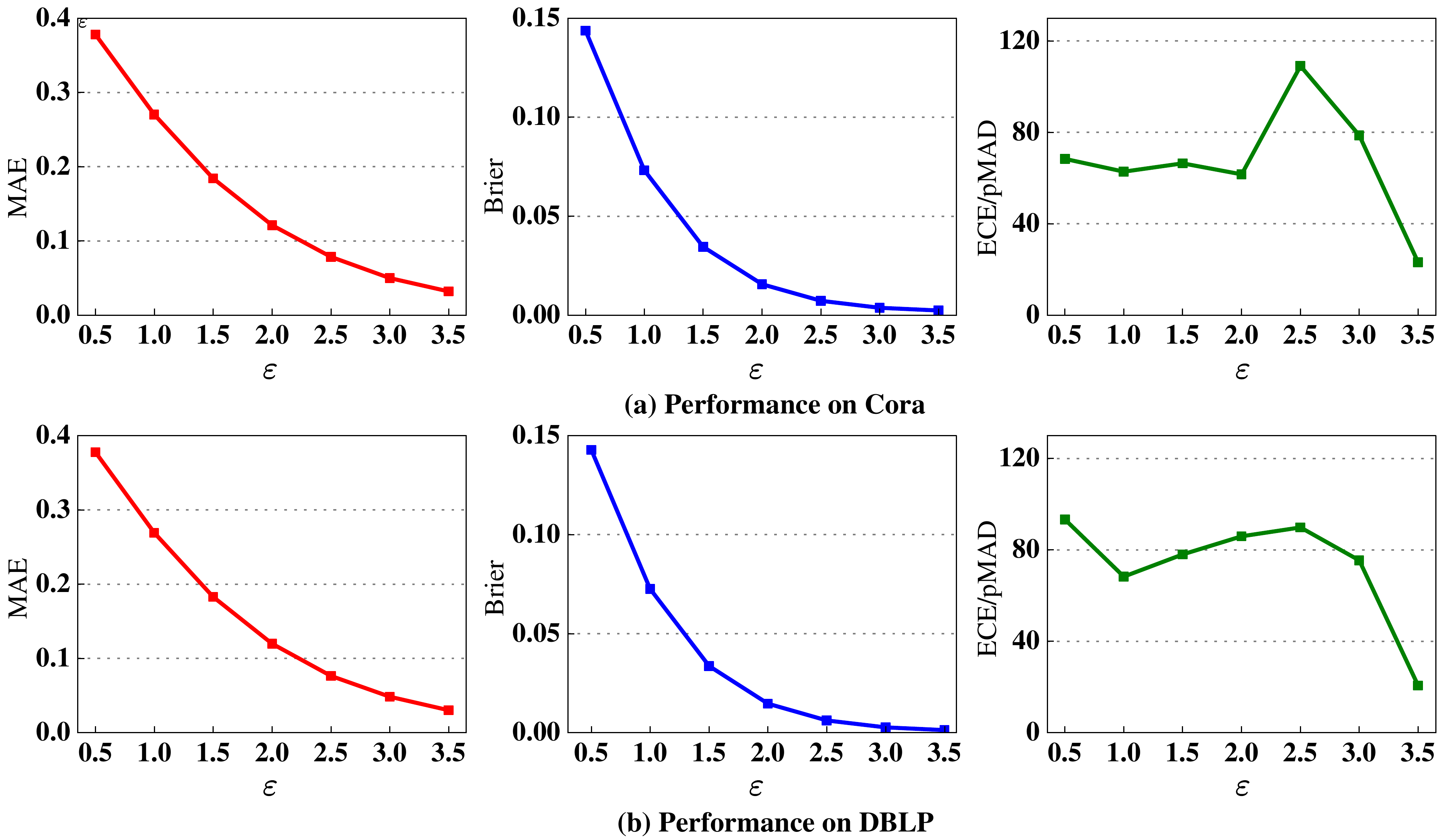}
    \caption{Performance of Probability Estimation Error}
    \label{err}
\end{figure}

We conducted a comprehensive evaluation of the probability estimation quality in our framework by comparing the estimated probability matrix against the ground-truth adjacency matrix. We selected two datasets, Cora and DBLP, which are representative in terms of accuracy trends. The evaluation was performed under varying privacy budgets $\epsilon$ ranging from 0.5 to 3.5, using three error metrics: MAE, Brier score, and the ECE/pMAD ratio. The experimental result is shown in Figure~\ref{err}. 

On the Cora dataset as shown in Figure~\ref{err}(a), all metrics follow the general improvement trend observed in the other datasets. The MAE reduces from 0.3779 to 0.0320, and the Brier score declines from 0.1437 to 0.0023, both demonstrating substantial enhancements exceeding 90\%. The ECE/pMAD ratio decreases from approximately 68.4 to 23.1, but exhibits the most pronounced fluctuations among the three datasets. The ratio shows an initial rapid decline, followed by a temporary increase around $\epsilon\in [2.5, 3.0]$, before settling at the final value. This non-monotonic behavior indicates that the calibration quality relative to dispersion may be influenced by dataset-specific characteristics and the complex interplay between noise reduction and probability estimation accuracy. Across all three datasets, the consistent improvement in MAE and Brier score confirms that our probability estimation method effectively leverages reduced noise, while the ECE/pMAD patterns reveal the nuanced relationship between calibration quality and estimate variability under different privacy constraints.

On the DBLP dataset in Figure~\ref{err}(b), the performance of our probability estimation is consistent with the trends observed on the other datasets. All metrics demonstrate significant performance improvement as the privacy budget increases. The MAE and Brier score exhibit a clear decreasing trend, with the latter nearing zero at higher privacy budgets, indicating enhanced probabilistic forecasting accuracy. The ECE/pMAD ratio also follows a generally declining but non-monotonic pattern, similar to the behavior observed on the Cora dataset. 

Despite these improvements in probability estimation quality, the final classification performance of EdgeRefine does not strongly depend on the absolute accuracy of the probabilities. For instance, even at low $\epsilon$ values where MAE and Brier scores are relatively high (e.g., $\epsilon=0.5$), our framework maintains competitive node classification accuracy, as shown in earlier experiments.  ECE/pMAD's non-monotonic behavior, which shows similar patterns across datasets, suggests that ECE/pMAD may serve as a more sensitive indicator of the underlying dynamics in probability estimation. Therefore, while high-quality probability estimation is beneficial, it is not the primary driver of denoising effectiveness; our framework's robustness stems from the integrated approach that balances multiple components. 

The expansion experiments also confirm that EdgeRefine maintains close classification performance over a large privacy budget span. Detailed experimental data are presented in  Appendix~\ref{ex-high}.

\subsubsection{Parameter Analysis (\textbf{RQ4})}
\label{par exp}

We conducted a comprehensive evaluation of our proposed framework on two widely-used citation networks, DBLP and Cora, to systematically assess its robustness and scalability. The experiments were performed under a strict privacy budget of $\epsilon=2.0$, employing GAT, GCN, and GIN. The models were trained and evaluated under a range of sampling rates $k$ (specifically, $k = 0.005, 0.01, 0.05, 0.1, 0.3, 0.5$) to examine the relationship between graph sparsification and model utility. The detailed results, encapsulating the test accuracy for each model-architecture-dataset combination, are comprehensively summarized in Table~\ref{tab:parameter_analysis}.

\begin{DIFnomarkup}

\begin{table}[htbp]
\small
\centering
\caption{Performance of EdgeRefine under Different Sampling Rates}
\label{tab:parameter_analysis}

\setlength{\tabcolsep}{4pt}
\begin{tabular}{ll|cccccc}
\hline
\multirow{2}{*}{\textbf{Dataset}} & \multirow{2}{*}{\textbf{Model}} & \multicolumn{6}{c}{\textbf{Sampling Rate $k$}} \\
\cline{3-8}

& & 0.005  & 0.01            & 0.05            & 0.1             & 0.3    & 0.5    \\
\hline
\multirow{3}{*}{DBLP} & GAT & 0.8153 & 0.7919 & \textbf{0.8350} & 0.8239 & 0.8116 & 0.7783 \\
                      & GCN & 0.8079 & 0.7833 & 0.7882 & \textbf{0.8140} & 0.8103 & 0.7833 \\
                      & GIN & 0.7746 & 0.7734 & \textbf{0.7943} & 0.7894 & 0.7660 & 0.7328 \\
\hline
\multirow{3}{*}{Cora} & GAT & 0.7495 & \textbf{0.7514} & 0.7477 & 0.7164 & 0.6759 & 0.6648 \\
                      & GCN & 0.7109 & \textbf{0.7182} & 0.7164 & 0.6998 & 0.7072 & 0.6777 \\
                      & GIN & 0.6740 & \textbf{0.7035} & 0.6943 & 0.6335 & 0.5948 & 0.5617 \\
\hline
\end{tabular}
\end{table}

\end{DIFnomarkup}

A pivotal and consistent observation from our analysis is the non-monotonic relationship between the sampling rate $k$ and the final utility of the models across both datasets and all three architectures. For the Cora dataset, the GAT architecture demonstrates an initial improvement in accuracy, rising from 0.7495 at $k=0.005$ to a peak of 0.7514 at $k=0.01$, after which it experiences a gradual decline as $k$ increases further, eventually reaching 0.6648 at $k=0.5$. A remarkably similar trend is observed for GCN on Cora, where accuracy increases from 0.7109 at $k=0.005$ to a maximum of 0.7182 at $k=0.01$, before decreasing to 0.6777 at $k=0.5$. Interestingly, the GCN model exhibits a minor performance recovery at $k=0.3$ (accuracy: 0.7072), hinting at a potentially different sensitivity profile compared to GAT. The GIN architecture on Cora also achieves its best performance (0.7035) at $k=0.01$, but appears more sensitive to higher $k$ values, experiencing a sharper decline in accuracy. On the larger and more complex DBLP dataset, the optimal sampling point varies. The GAT model peaks at a higher $k=0.05$ with an accuracy of 0.8350, suggesting it might benefit from a denser, albeit noisier, graph structure on this dataset. Conversely, the GCN model on DBLP achieves its best performance (0.8140) at $k=0.1$ , while GIN, similar to its behavior on Cora, performs best (0.7943) at $k=0.05$ . This variation underscores that the optimal $k$ is not universal but is contingent upon the specific dataset characteristics and the intrinsic properties of the GNN architecture employed.

The trend is explained by $k$'s dual effect on graph structure. Lower $k$ values enhance edge credibility by selectively preserving high-probability connections, boosting Signal-to-Noise Ratio (SNR), but may oversparsify the graph. Higher $k$ values improve connectivity at the cost of introducing noisy edges, reducing SNR. Optimal performance balances sufficient connectivity with minimal noise.

\subsubsection{Ablation Study (\textbf{RQ5})}
\label{abl exp}

\begin{figure}[htbp]
    \centering
    \includegraphics[width=\columnwidth]{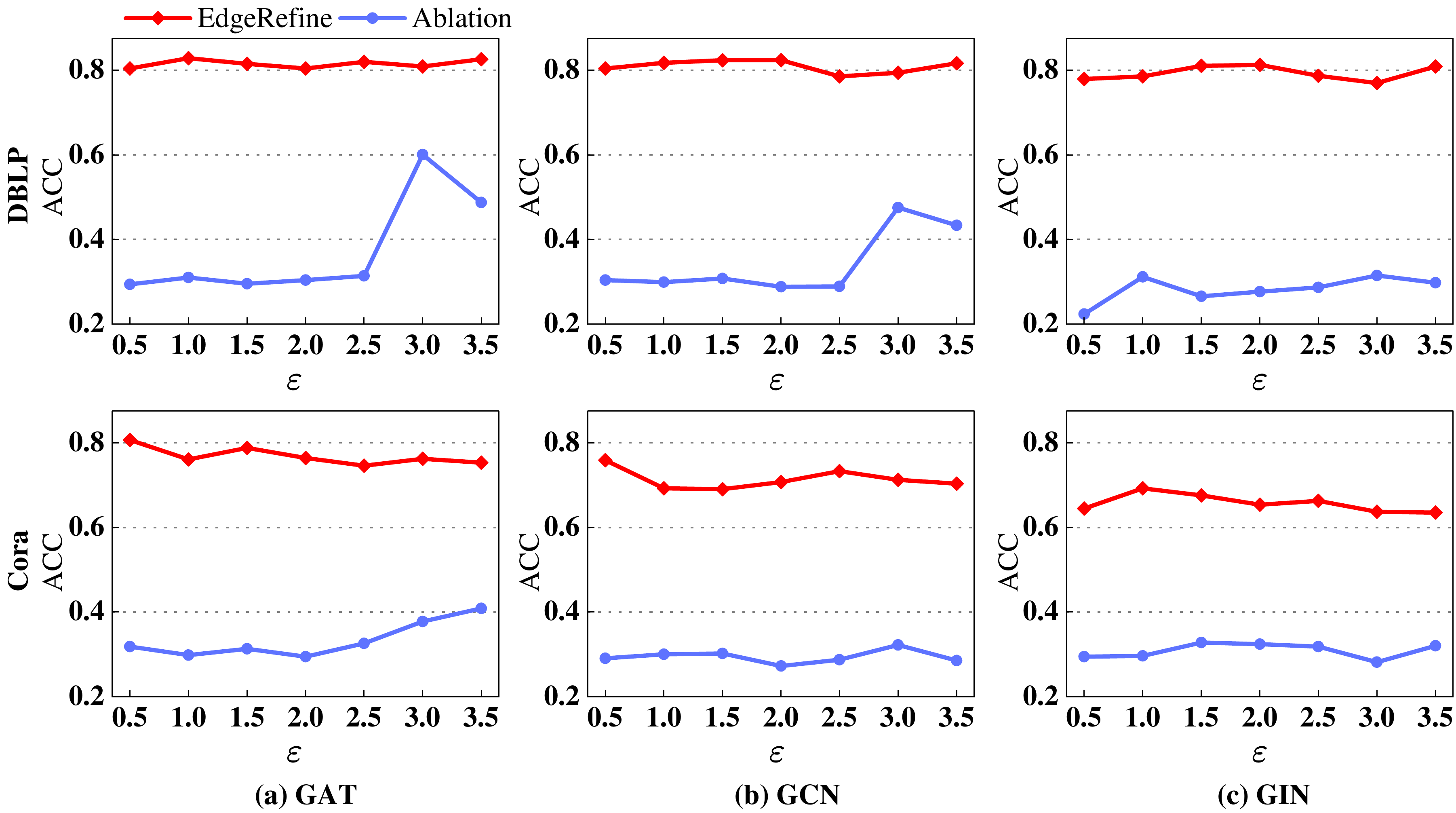}
    \caption{Ablation Study}
    \label{abl}
\end{figure}

We conducted a comprehensive ablation study to quantitatively evaluate the contribution of the proposed privacy-preserving components in our EdgeRefine framework. The study systematically compares the performance of our full framework against an ablated version, where critical mechanisms including Jaccard-based probability estimation and deterministic sampling modules were removed, retaining only the basic randomized response perturbation mechanism. This design allows us to isolate the effects of individual components and assess their importance in maintaining utility under differential privacy constraints. All experiments were performed on both Cora and DBLP datasets across three representative GNN architectures (GAT, GCN, GIN) under varying privacy budgets $\epsilon$ ranging from 0.5 to 3.5, with performance measured using node classification accuracy. The experimental results are shown in Figure~\ref{abl}. The reported results are averaged over three runs, which accounts for the slight differences from the results shown in Figure~\ref{exp}. 

The DBLP dataset results demonstrate the performance advantages of our framework. EdgeRefine maintains accuracy around 0.8 across the privacy budget spectrum. In contrast, the accuracy of the ablated version remains around 0.3 for privacy budgets with $\epsilon<2.5$. Although its performance shows a slight improvement at higher budgets ($\epsilon \geq 2.5$) in Figure~\ref{abl}(a), it does not exceed 0.6 under the GAT architecture, which is more than 20\% lower than our results. This indicates that the ablated version cannot fully leverage the relaxed privacy constraints to close the performance gap.

On the Cora dataset in Figure~\ref{abl}, EdgeRefine achieves an average accuracy above 0.76 with GAT across all evaluated privacy budgets, while consistently outperforming the ablated variant across all three GNN architectures. For example, under the GAT architecture with a privacy budget of \( \epsilon=0.5 \),  our framework achieves an accuracy of 0.8066, significantly outperforming the ablated version's 0.3005. The average improvement of 0.4398 absolute accuracy points. Even when the ablated version yields minor gains at higher privacy budgets, it fails to close the performance gap, while our EdgeRefine under GIN architecture in Figure~\ref{abl}(c) maintains a 14.4\% relative improvement over the ablated one.

In addition, we also conducted experiments to evaluate our deterministic sampling method separately. Specifically, we use the probability matrix obtained from the Jaccard probability estimation section to compare our deterministic sampling and the commonly used effective method in probability estimation, top-k sampling. The experimental results show that the sampling method under sparsity constraints and true false ratio constraints is better than sampling sorted by probability, which proves the effectiveness of our proposed EdgeRefine. We further performed a comprehensive comparison of different probability estimation and calibration methods, with results confirming the superiority of the Jaccard+histogram combination. Detailed results of these experiments are provided in Appendix~\ref{ex-ab} and Appendix~\ref{ex-mt}.

\subsubsection{Graph Classification Results (\textbf{RQ6})}\label{sec-graphclass}
We further conducted experiments to evaluate the performance of EdgeRefine on the graph classification task. Table \ref{tab:graph_classification} presents the average classification accuracy of GAT, GCN, and GIN models on perturbed graphs under different $\epsilon$ values on the MUTAG~\cite{mutag} dataset, compared with the original noise-free graph from client.

\begin{table}[htbp]
\small
\centering
\caption{Graph Classification Accuracy under Varying $\epsilon$ Values}
\setlength{\tabcolsep}{2.5pt}
\begin{tabular*}{\columnwidth}{
@{\hspace{4pt}}
l
@{\hspace{4pt}\extracolsep{\fill}}  % 这里加6pt固定间距，拉开Model和Noise‑free
cccccccc
@{\hspace{4pt}}  % 最右侧留白
}
\toprule
\textbf{Model} & \textbf{Noise-free}  & \textbf{ 0.5 } & \textbf{ 1.0 } & \textbf{ 1.5 } & \textbf{ 2.0 } & \textbf{ 2.5 } & \textbf{ 3.0 } & \textbf{ 3.5 }\\
\midrule
GAT & 0.684 & 0.589 & 0.579 & 0.642 & 0.632 & 0.663 & 0.642 & 0.737 \\
GCN & 0.611 & 0.642 & 0.642 & 0.589 & 0.589 & 0.611 & 0.642 & 0.674 \\
GIN & 0.747 & 0.716 & 0.695 & 0.684 & 0.684 & 0.758 & 0.747 & 0.747 \\
\bottomrule
\end{tabular*}
\label{tab:graph_classification}
\end{table}

From Table \ref{tab:graph_classification}, we observe that the classification performance remains competitive with the noise-free baseline even under substantial privacy constraints ($\epsilon \geq 2.5$). For GIN, the accuracy consistently approaches the noise-free baseline, with $\epsilon = 2.5$ achieving 0.758, surpassing the noise-free baseline. The GAT model also demonstrates stable performance, particularly with $\epsilon = 3.5$ yielding 0.737, which is comparable to the noise-free baseline. In general, our perturbation mechanism preserves essential structural information, enabling GNNs to maintain effective performance for downstream graph classification tasks while providing rigorous privacy guarantees.

Although the link prediction task fundamentally conflicts with the goal of protecting edge privacy, we still conducted small-scale experiments. The results indicate that the link prediction task performed poorly, consistent with our expectations, as detailed in  Appendix~\ref{ex-lp}.

\subsubsection{Anti Attack Performance (\textbf{RQ7})}
\label{antiattack}
The resistance to inference attacks is a critical metric for evaluating the practical privacy guarantees of graph perturbation mechanisms. We employ the GRAND~\cite{grand} attack, a state-of-the-art graph reconstruction attack, to assess the vulnerability of our perturbed graphs. The attack attempts to reconstruct the original graph structure from the perturbed adjacency matrix $\hat{A}$. 
\begin{table}[htbp]
\centering
\caption{RAE (Resistance to GRAND Attack) under Different Privacy Budgets}
\label{tab:grand_attack}
\small
\begin{tabular*}{\columnwidth}{@{\hspace{4pt}} l @{\hspace{6pt}\extracolsep{\fill}} cccccccc @{\hspace{3pt}}}
\toprule
\textbf{$\epsilon$} & \textbf{0.5} & \textbf{1.0} & \textbf{1.5} & \textbf{2.0} & \textbf{2.5} & \textbf{3.0} & \textbf{3.5} \\
\midrule
Cora & 1.9977 & 1.9456 & 1.9373 & 1.9848 & 1.9689 & 1.9485 & 1.9488 \\
AMAP & 1.6134 & 1.6127 & 1.6107 & 1.5407 & 1.4599 & 1.3244 & 1.1425 \\
\bottomrule
\end{tabular*}
\end{table}

Table~\ref{tab:grand_attack} presents the resistance of our perturbation mechanism against the GRAND attack across different privacy budgets on the Cora and AMAP datasets. The RAE values are consistently above 1.9 for the Cora dataset and above 1.1 for the AMAP dataset across all privacy budgets, indicating strong resistance to graph reconstruction attacks. Notably, on the Cora dataset, the RAE remains high (averaging 1.962) regardless of the privacy budget, demonstrating that even with relatively weak privacy protection ($\epsilon=3.5$), the perturbed graph structure effectively obfuscates the original topology. For the larger AMAP dataset, although the RAE decreases as $\epsilon$ increases (from 1.6134 at $\epsilon=0.5$ to 1.1425 at $\epsilon=3.5$), it remains on average at 1.472, indicating that the attack consistently fails to accurately reconstruct the original graph. This robustness  stems from the random edge perturbation mechanism, which introduces sufficient uncertainty in the graph structure to thwart inference attacks while preserving essential topological properties for downstream tasks.

Our results show that our method maintains high RAE values, especially on Cora, where the RAE is almost constant and close to 2, indicating that the attacker's recovery ability is severely hindered.

\textbf{Overall Result:} These results collectively demonstrate that the proposed components essentially contribute to maintaining utility under privacy constraints. The integration of Jaccard-based probability estimation with deterministic sampling creates a synergistic effect that enables effective edge selection while preserving structural properties. The comprehensive experimental results further demonstrate that our proposed method effectively addresses the three key technical challenges identified earlier. Firstly, the probability error of our estimation is at a low level across all privacy budget intervals and rapidly decreases as the privacy budget increases. This indicates that the probability we calculat can effectively identify trustworthy edges and solve \textbf{TC1}. Secondly, the node classification accuracy of our framework remains close to the noise-free baseline throughout the entire privacy budget range, showing a very stable trend. This indicates that even when the probability error is relatively high, we can still maintain utility without relying solely on the quality of probability. These performances strongly demonstrate that we have solved \textbf{TC2}. Finally, our deterministic sampling algorithm ensures absolute stable sparsity, which brings significant advantages in terms of time overhead. Our framework outperforms almost all baseline schemes in terms of sparsity and time cost metrics across all privacy budgets. Therefore, our framework can be considered as the best solution to \textbf{TC3}.

\section{Conclusion}
To address the challenge of balancing edge-level differential privacy with utility, this paper presents EdgeRefine, a practical framework for privacy-preserving graph learning. EdgeRefine applies Jaccard similarity and deterministic sampling to preserve graph structure while maintaining privacy guarantees. The experiments show that EdgeRefine delivers consistent performance across four diverse datasets and three GNN architectures. It achieves stable performance comparable to the noise-free baseline, with an accuracy of approximately 0.9 on ACM dataset. 
The graph classification task and attack experiments also demonstrate EdgeRefine's superior performance on other downstream tasks as well as its strong privacy-preserving capability. %In most combinations of dataset architecture, EdgeRefine is superior to state-of-the-art baselines in all aspects, and the remaining parts also maintain an overall advantage.

The framework establishes a foundation for reliable privacy-preserving graph analysis and is amenable to extension for dynamic network settings.

\section*{Acknowledgments}
This paper used deepseek for grammar checking and minor language polishing. This work is supported in part by the National Natural Science Foundation of China under Grant U23A20300; in part by the ``Pioneer” and ``Leading Goose” R\&D Program of Zhejiang under Grant No. 2026C01020; in part by the Concept Verification Funding of Hangzhou Institute of Technology of Xidian University under Grant GNYZ2024XX007; in part by the Fundamental Research Funds for the Central Universities under Grant QTZX26067; and in part by the 111 Project under Grant B16037.

\bibliography{ref} 

\appendix
\section*{Open Science}

To support reproducibility and facilitate further research in privacy-preserving graph machine learning, we provide a public repository for the preprint version of this work at \url{https://github.com/Liumz-1/EdgeRefine-open}. 
\begin{itemize}
\item \textbf{Source Code:} The full implementation of the EdgeRefine framework, written in Python using PyTorch and PyTorch Geometric. The code is documented to facilitate use and extension by the research community.
\item \textbf{Processed Datasets:} The standardized versions of the four benchmark datasets (ACM, DBLP, AMAP, and Cora) used in our experiments, together with scripts for downloading and preprocessing the raw data into the format used in this study.
\item \textbf{Baseline Implementations:} Our implementations of the privacy-preserving baseline methods used in the comparative experiments, configured with the parameters reported in the paper.
\item \textbf{Raw Experimental Data:} The key outputs from our experiments, including final evaluation metrics such as accuracy under different privacy budgets $\epsilon$ and other relevant logs. These data provide the exact numbers behind the figures and analysis presented in the paper.
\item \textbf{Documentation:} A comprehensive \texttt{README.md} file describing the experimental setup, dependency installation, data preparation, and commands required to reproduce the key results and figures.
\end{itemize}

The GitHub repository is intended to provide convenient access to the preprint-version artifacts and may be updated or relocated in future releases.

\begin{DIFnomarkup}

\section{Theoretical Analysis of Security and Utility}
\label{the}
Our EdgeRefine framework establishes a theoretically-grounded framework to privacy-preserving graph analysis through careful integration of differential privacy mechanisms with graph structural properties. The core innovation lies in our strategic use of Jaccard similarity as a robust estimator that naturally resists noise, combined with histogram binning and controlled sampling to balance privacy and utility. Below, we present a structured theoretical analysis of key components.

\textbf{Robustness of Jaccard Similarity to Noise:}
The Jaccard similarity measure proves particularly effective because it captures meaningful structural relationships that persist even after edge-level perturbations. While individual edges may be flipped during randomization, the overall neighborhood structure exhibits remarkable stability. This arises from Jaccard similarity's focus on relative overlap rather than absolute edge counts, making it inherently resilient to random perturbations. Quantitatively, consider the effect of a single edge flip: for node degree $d_i$, the change is linear:

\[
\Delta d_i = 1
\]
whereas for Jaccard similarity $J(i,j) = \frac{|N(i)\cap N(j)|}{|N(i)\cup N(j)|}$, the change is bounded by:
\[
|\Delta J(i,j)| \leq \frac{1}{|N(i)\cup N(j)|}
\]

This indicates that the impact of noise on the Jaccard coefficient is relatively stable.

\textbf{Theoretical Advantages of Histogram Binning:}
Our histogram binning approach provides superior robustness compared to parametric methods such as ReLU-based neural networks. For a histogram with $B$ bins, the mean squared error decays as:
\[
\mathbb{E}[(\hat{p}_b - p_b)^2] = O\left(\frac{1}{n_b} + \frac{1}{B^2}\right)
\]
In contrast, parametric models suffer from approximation error:
\[
\mathbb{E}[(\hat{p}_{NN} - p)^2] = O\left(\frac{V}{n} + \epsilon_{approx}\right)
\]
where $N$ is the number of nodes, $n=N(N-1)/2$ is the total
number of unordered node-pair samples, and $n_b$ is the number
of samples assigned to bin $b$.
 With $V$ denoting VC-dimension and $\epsilon_{approx}$ the approximation error. Histogram estimators avoid $\epsilon_{approx}$ entirely and are less prone to overfitting. Additionally, the calibration error satisfies:
\[
\mathbb{E}[|p_b - \hat{p}_b|] \leq \sqrt{\frac{\log B}{2n}}
\]
outperforming neural networks that may miscalibrate near decision boundaries. This data-driven approach adapts to diverse graph types without manual tuning.

\textbf{Signal-to-Noise Ratio Enhancement via Controlled Sampling:}
The strategic reduction in graph density through controlled sampling enhances both privacy protection and graph quality. By focusing on high-probability edges, we amplify the Signal-to-Noise Ratio (SNR). Define precision $P(k)$ as the proportion of true edges among sampled edges at rate $k$; then:
\[
\text{SNR}(k) = \frac{P(k)}{1 - P(k)}
\]
As $k$ decreases, the probability threshold $\tau_k$ (the $(1-k)$-th quantile) increases, leading to higher precision $P(k) \geq \tau_k$ when estimates are well-calibrated. Thus, SNR increases monotonically with decreasing $k$, improving the reliability of the denoised graph. This selective preservation is crucial for GNNs, which depend on clean structural patterns for message passing.

\textbf{Synergistic Integration for Balanced Privacy-Utility:}
The theoretical strength of EdgeRefine emerges from the synergistic combination of differential privacy guarantees and deterministic probability estimation. The DP framework provides mathematical certainty about privacy protection, while Jaccard-based estimation and histogram binning ensure practical utility preservation. The post-processing properties of DP ensure strict privacy protection even after probability estimation and sampling. This integration addresses both theoretical requirements (e.g., bounded sensitivity) and practical considerations (e.g., scalability across graph types), creating a balanced solution for real-world graph analysis.

In summary, our theoretical analysis demonstrates that EdgeRefine’s components collaborate well to yield an optimal privacy-utility balance. The Jaccard similarity's noise resilience, histogram binning's error bounds, and controlled sampling's SNR enhancement collectively provide a foundation for reliable and effective privacy-preserving graph learning. This theoretical grounding supports the empirical results observed in our experiments.
\end{DIFnomarkup}

\section{Graph Neural Networks}
\label{gn}
GNNs represent a class of deep learning architectures specifically designed for processing graph-structured data~\cite{graphdata}. The fundamental concept behind GNNs is the message-passing mechanism, where each node iteratively aggregates information from its neighbors to update its own representation. This process can be formally described by the general message-passing framework:

\begin{equation}
h_v^{(k)} = \phi\left(h_v^{(k-1)}, \psi\left(\{h_u^{(k-1)} : u \in \mathcal{N}(v)\}\right)\right)
\label{eq:gnn_general}
\end{equation}

where $h_v^{(k)}$ denotes the representation of node $v$ at layer $k$, $\mathcal{N}(v)$ represents the neighborhood of node $v$, $\phi$ is the update function, and $\psi$ is the aggregation function. Among various GNN architectures, three prominent models have demonstrated exceptional performance across diverse graph learning tasks.

\textbf{Graph Attention Network (GAT)~\cite{GAT}} introduces attention mechanisms to assign learnable weights to neighbors:
\begin{equation}
h_v^{(l+1)} = \sigma\left(\sum_{u\in\mathcal{N}(v)}\alpha_{vu}W^{(l)}h_u^{(l)}\right)
\label{eq:gat}
\end{equation}
where $\alpha_{vu}$ denotes the attention coefficient between nodes $v$ and $u$. GAT enables nodes to attend to their neighbors with varying importance, enhancing model expressiveness and interpretability.

\textbf{Graph Convolutional Network (GCN)~\cite{GCN}} employs spectral graph convolution with a first-order approximation, defined as:
\begin{equation}
H^{(l+1)} = \sigma\left(\tilde{D}^{-1/2}\tilde{A}\tilde{D}^{-1/2}H^{(l)}W^{(l)}\right)
\label{eq:gcn}
\end{equation}
where $\tilde{A} = A + I$ is the adjacency matrix with self-connections, $\tilde{D}$ is the degree matrix of $\tilde{A}$, and $W^{(l)}$ represents the trainable weight matrix. GCN provides an efficient approach for semi-supervised learning on graph-structured data by leveraging spectral graph theory.

\textbf{Graph Isomorphism Network (GIN)~\cite{GIN}} provides theoretically powerful expressiveness by leveraging the Weisfeiler-Lehman test:
\begin{equation}
h_v^{(k)} = \text{MLP}^{(k)}\left((1+\epsilon)h_v^{(k-1)} + \sum_{u\in\mathcal{N}(v)}h_u^{(k-1)}\right)
\label{eq:gin}
\end{equation}
GIN is provably as powerful as the Weisfeiler-Lehman graph isomorphism test, making it capable of capturing subtle structural differences between graphs.

These architectures, while employing different aggregation and update strategies, share the common objective of effectively capturing both structural information and node features for various downstream tasks including node classification, link prediction, and graph classification~\cite{GNN}.
\section{Expansion Experiment}
\label{ex-exp}

\subsection{Standard Deviation of Multiple Experiments}
\label{ex-avg}

To eliminate experimental randomness and further evaluate the stability of model performance, we further analyze the standard deviation of experimental results.
For each privacy budget $\epsilon  \in \{0.5, 1.0, ..., 3.5\}$, we ran the experiment five times and calculated the standard deviation of accuracy. Given the large volume of results, Table~\ref{tab:avg-std} reports the average of these seven standard deviations, reflecting the overall stability of our method across different privacy levels.

\begin{table}[htbp]
\caption{Average Standard Deviation Across Privacy Budgets ($\epsilon=0.5$ to $3.5$)}
\label{tab:avg-std}
\centering
\footnotesize
\setlength{\tabcolsep}{1.55pt}

\begin{tabular*}{\columnwidth}{@{\hspace{1pt}} p{0.08\columnwidth} @{\extracolsep{\fill}} l *{6}{c} @{\hspace{1pt}}}

\toprule
Arch. & Dataset & \textbf{EdgeRefine} & Blink-hard & Blink-hybrid & DPRR & LAPGRAPH & LDPGen \\
\midrule
\multirow{4}{*}{GAT} & ACM & \underline{0.0078} & \textbf{0.0185} & \textbf{0.0185} & 0.0529 & 0.0202 & 0.0391 \\ 
 & AMAP & 0.0155 & \underline{0.0102} & \textbf{0.0116} & 0.0674 & 0.1391 & 0.0724 \\ 
 & Cora & \underline{0.0147} & \textbf{0.0179} & \textbf{0.0179} & 0.0186 & 0.0187 & 0.0180 \\ 
 & DBLP & \underline{0.0112} & \textbf{0.0130} & \textbf{0.0130} & 0.0248 & 0.0150 & 0.0172 \\ \midrule
\multirow{4}{*}{GCN} & ACM & \underline{0.0111} & 0.0214 & 0.0203 & \textbf{0.0134} & 0.0394 & 0.0327 \\ 
 & AMAP & 0.1301 & 0.0674 & 0.0714 & \underline{0.0089} & 0.0562 & \textbf{0.0311} \\ 
 & Cora & 0.0217 & 0.0261 & 0.0226 & 0.0235 & \textbf{0.0185} & \underline{0.0176} \\ 
 & DBLP & \underline{0.0100} & 0.0142 & \textbf{0.0110} & 0.0154 & 0.0160 & 0.0132 \\ \midrule
\multirow{4}{*}{GIN} & ACM & \underline{0.0126} & 0.0818 & 0.0818 & \textbf{0.0168} & 0.0476 & 0.0525 \\ 
 & AMAP & 0.0944 & 0.1544 & 0.1303 & \underline{0.0089} & \textbf{0.0111} & 0.0157 \\ 
 & Cora & \textbf{0.0172} & 0.0298 & 0.0298 & \underline{0.0161} & 0.0255 & 0.0208 \\ 
 & DBLP & \underline{0.0120} & 0.0189 & 0.0189 & 0.0176 & 0.0138 & \textbf{0.0130} \\ 
\midrule
\multicolumn{2}{@{}l}{Average} & 0.0299 & 0.0395 & 0.0373 & 0.0237 & 0.0351 & 0.0286 \\ 
\bottomrule
\end{tabular*}

\begin{tablenotes}
    \small
    \item \textbf{Notes:} In each row, the best result is \underline{underlined}, while the second-best result is highlighted in \textbf{bold}.
\end{tablenotes}

\end{table}

Several observations emerge from the table. Overall, EdgeRefine achieves low averaged standard deviations across most dataset--architecture combinations, particularly on ACM and DBLP, where variance remains minimal across all privacy levels. This confirms that our framework delivers stable performance in the vast majority of cases.

That said, AMAP does introduce isolated instability, though its severity varies by architecture. From the average standard deviations (averaged over all privacy budgets), two combinations emerge as outliers: GCN-AMAP (0.1301) and GIN-AMAP (0.0944). Importantly, their underlying stability profiles differ significantly. Specifically, GCN-AMAP's larger variance is distributed relatively uniformly across privacy budgets, whereas GIN-AMAP, despite its lower average standard deviation, exhibits sporadic extreme deviations—most notably a sharp spike at $\epsilon=3.0$.

Other schemes reflect similar disparities: Blink-hard and Blink-hybrid on GIN-AMAP reach 0.1544 and 0.1303, respectively, and LAPGRAPH on GAT-AMAP spikes to 0.1391.

These elevated deviations are confined to specific model--dataset pairings. The cross-architecture pattern suggests that AMAP's sparser or more heterogeneous topology can amplify sensitivity for certain combinations, yet EdgeRefine preserves strong overall stability on most benchmarks. In the following subsection, we conduct a targeted $t$-test on these high-variance points to determine whether the observed fluctuations stem from the privacy mechanism or from inherent model--dataset sensitivity.

\subsection{Special Case Analysis}
\label{ex-uno}

The performance curves in Figure~\ref{exp} reveal an intriguing anomaly on the AMAP dataset with the GIN architecture: the EdgeRefine model under differential privacy occasionally matches or even surpasses the noise-free baseline. This contradicts the conventional expectation that privacy-preserving mechanisms strictly degrade utility. To investigate this phenomenon rigorously, we conducted an additional controlled experiment.

When initialized with the same random seed, GIN deterministically produces the exact same accuracy across repeated runs on this dataset. Consequently, a single-seed comparison would offer no statistical basis for inference. We therefore ran both the noise-free baseline and EdgeRefine with $\epsilon=3.0$ under 8 different initialization conditions (seeds 47 through 54), isolating the effect of the privacy mechanism from the variance induced by different random seeds.

The noise-free baseline exhibits extreme instability,  as corroborated by Table~\ref{tab:avg-std} in Appendix ~\ref{ex-avg}: AMAP consistently induces higher variance across all three architectures (GAT, GCN, and GIN) compared to ACM, Cora, and DBLP. This cross-architecture pattern confirms that the observed instability is not specific to GIN, but stems from structural properties of AMAP itself---likely its sparser or more heterogeneous topology---which fundamentally amplify the sensitivity of all models to experimental variation.

Table~\ref{tab:ttest_simple} reports the independent two-sample $t$-test on the 8 runs. The test yields $t = 1.267$ with a two-sided $p$-value of $0.226$. We fail to reject the null hypothesis: there is no statistically significant difference between EdgeRefine at $\epsilon=3.0$ and the noise-free baseline.

\begin{table}[htbp]
\centering
\caption{Performance Comparison: Noise-Free Baseline vs. EdgeRefine ($\epsilon=3.0$) on GIN-AMAP}
\label{tab:ttest_simple}
\small
\begin{tabular*}{\columnwidth}{@{\hspace{4pt}} l @{\extracolsep{\fill}} c c c c @{\hspace{4pt}}}
\toprule
\textbf{Metric} & \textbf{Number of runs} & \textbf{Mean} & \textbf{Std} & \textbf{$p$-value} \\
\midrule
Noise-Free Baseline & 8 & 0.3407 & 0.1352 & \multirow{2}{*}{0.226} \\
EdgeRefine ($\epsilon=3.0$) & 8 & 0.4454 & 0.1907 & \\
\bottomrule
\end{tabular*}
\end{table}

This finding provides a definitive explanation for the apparent anomaly. The fluctuations previously perceived as privacy-induced are, in fact, manifestations of the GIN model's inherent instability on the AMAP dataset. When a base model exhibits such high variance, the additional perturbation from differential privacy at $\epsilon=3.0$ is statistically indistinguishable from the  noise-free baseline. In other words, the DP mechanism neither meaningfully degrades nor improves performance relative to the already erratic baseline; the two distributions are statistically equivalent.

For the GIN-AMAP case, the takeaway is clear: EdgeRefine preserves utility at $\epsilon=3.0$ at a level statistically on par with the noise-free baseline, and any local ``jumps'' in the privacy budget curve reflect the underlying model's volatility rather than the privacy mechanism's behavior.

\subsection{Analysis of the Relationship between Density and Accuracy}
\label{ex-de}

\begin{figure}[htbp]
    \centering
    \includegraphics[width=\columnwidth]{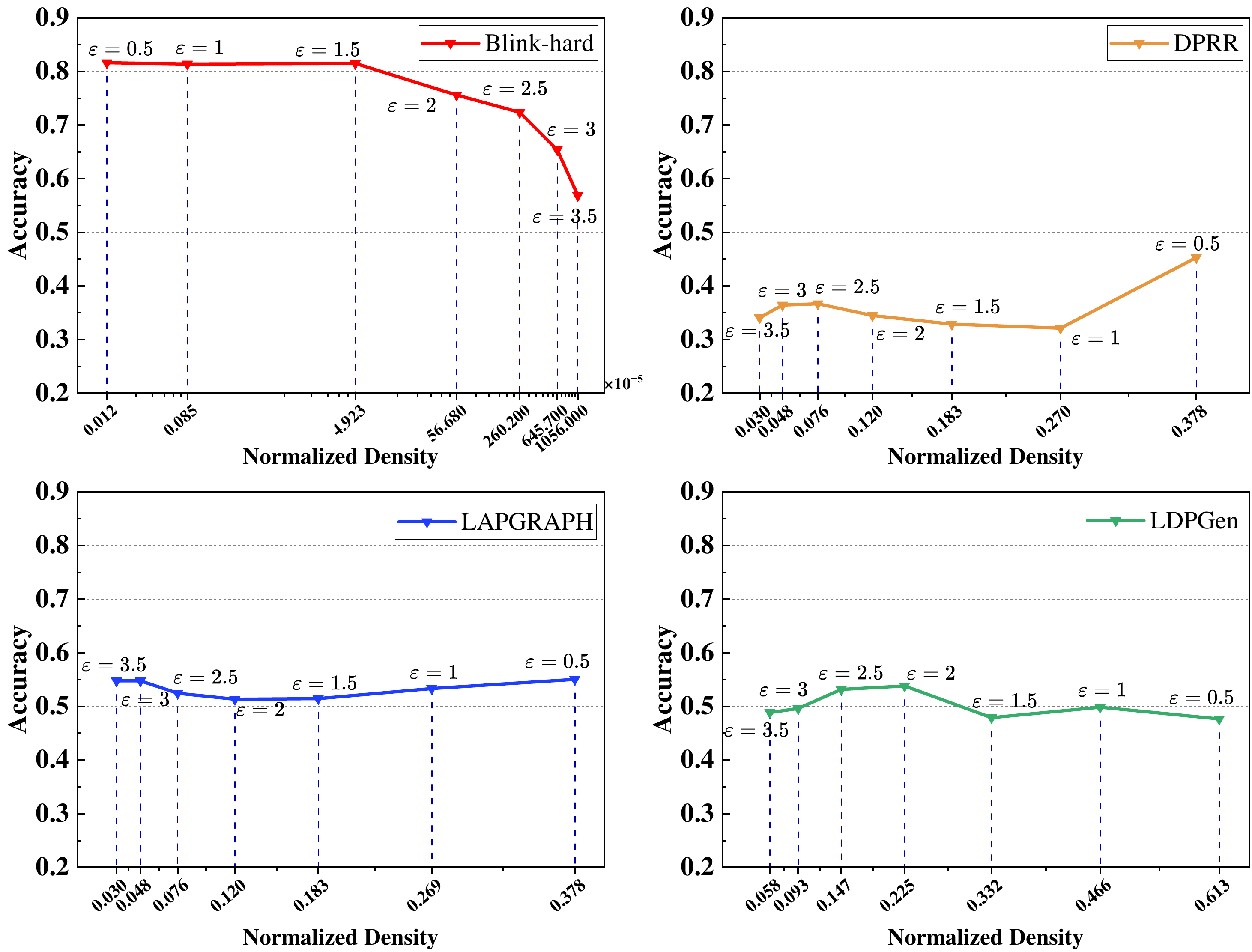}
    \caption{Baseline Density and Accuracy}
    \label{den}
\end{figure}

To show the relationship between density and accuracy, Figure~\ref{den} presents a detailed analysis of the GAT-DBLP results from Table~\ref{tab:density_values_transposed} (Section~\ref{ex-sp} ). Each density corresponds to a privacy budget from 0.5 to 3.5. It is worth mentioning that due to the different processing methods for noise edges in different schemes, their graph densities vary greatly. It can be seen that Blink-hard exhibits a significant negative correlation, which primarily stems from its sampling approach that causes dramatic variations in graph sparsity across different privacy budgets. In contrast, other baseline methods maintain relatively stable graph sparsity levels, typically remaining in lower/higher ranges that closely resemble Blink-hard's sparsity at privacy budget 3.5, where their accuracy performance is consistently low. Notably, our proposed scheme maintains constant sparsity throughout all privacy budgets, matching the sparsity magnitude of Blink-hard at privacy budget 1.0 while achieving comparable accuracy levels.

To complement the visualization above, we further quantified the correlation between density and accuracy across different schemes, as reported in Table~\ref{tab:density_accuracy_correlation}. Two standard statistical correlation measures are applied: Pearson's correlation coefficient (yielding a value $r$) and Spearman's rank correlation coefficient (yielding a value $\rho$). Pearson's $r$ quantifies the strength and direction of a \textit{linear} relationship between two continuous variables, ranging from $-1$ to $+1$. Spearman's $\rho$ assesses the strength and direction of a \textit{monotonic} relationship based on the rank order of the data, also ranging from $-1$ to $+1$, and offers robustness against outliers. Both coefficients are reported alongside corresponding \textit{p}-values. 

As shown in Table~\ref{tab:density_accuracy_correlation}, at the aggregate level across all methods, density exhibits a statistically significant negative correlation with accuracy (Pearson $r=-0.516$, $p=1.53\times10^{-3}$; Spearman $\rho=-0.763$, $p\approx10^{-7}$), indicating that greater structural perturbation systematically reduces GAT node classification accuracy. This trend is particularly pronounced within Blink-hard ($r=-0.981$): as $\epsilon$ increases, its normalized density rises from the $10^{-7}$ to the $10^{-2}$ order of magnitude, with accuracy dropping correspondingly from 0.817 to 0.569. However, the internal correlation coefficients for LDPGen, DPRR, and LAPGRAPH do not reach statistical significance, suggesting that the mapping between density metrics and accuracy in these methods may be dominated by other factors such as differences in perturbation mechanisms.

\begin{table}[htbp]
\centering
\caption{Correlation Between Normalized Density and Accuracy}
\label{tab:density_accuracy_correlation}
\small
\begin{tabular*}{\columnwidth}{@{\hspace{4pt}} l @{\extracolsep{\fill}} cccc @{\hspace{4pt}}}
\toprule
\multirow{2}{*}{\textbf{Method}} & \multicolumn{2}{c}{\textbf{Pearson}} & \multicolumn{2}{c}{\textbf{Spearman}} \\
\cline{2-3} \cline{4-5}
 & $r$ & $p$-value & $\rho$ & $p$-value \\
\midrule
EdgeRefine & --- & --- & --- & --- \\
Blink-hard & $\mathbf{-0.981}$ & $\mathbf{9.60 \times 10^{-5}}$ & $\mathbf{-0.964}$ & $\mathbf{4.54 \times 10^{-4}}$ \\
LDPGen & $-0.408$ & $0.363$ & $-0.286$ & $0.535$ \\
DPRR & $+0.510$ & $0.243$ & $0.000$ & $1.000$ \\
LAPGRAPH & $+0.105$ & $0.823$ & $+0.018$ & $0.969$ \\
\midrule
\textbf{Overall} & $\mathbf{-0.516}$ & $\mathbf{1.53 \times 10^{-3}}$ & $\mathbf{-0.763}$ & $\mathbf{9.99 \times 10^{-8}}$ \\
\bottomrule
\end{tabular*}%

\begin{tablenotes}
\small
\item \textbf{Notes:}  EdgeRefine exhibits zero variance in density across all $\epsilon$, rendering correlation undefined. Bold values indicate statistically significant correlations ($p < 0.05$). 
\end{tablenotes}

\end{table}

\subsection{ Stability under High Privacy Budgets}
\label{ex-high}
\begin{table}[htbp]
\centering
\small
\caption{Performance of EdgeRefine at High Privacy Budgets}
\label{tab:high_epsilon}

\begin{tabularx}{\columnwidth}{
  @{\hspace{3pt}}
  >{\centering\arraybackslash}p{1cm}
  | *{3}{X}
  | *{3}{X}
  @{\hspace{3pt}}
}
\hline
\multirow{2}{*}{\textbf{$\epsilon$}} & \multicolumn{3}{c|}{\textbf{DBLP}} & \multicolumn{3}{c}{\textbf{Cora}} \\
\cline{2-7}
 & GAT & GCN & GIN & GAT & GCN & GIN \\
\hline
0.5 & 0.8042 & 0.8042 & 0.7796 & 0.8066 & 0.7587 & 0.6446 \\
1.0 & 0.8288 & 0.8177 & 0.7857 & 0.7606 & 0.6924 & 0.6924 \\
1.5 & 0.8153 & 0.8239 & 0.8103 & 0.7882 & 0.6906 & 0.6759 \\
2.0 & 0.8042 & 0.8239 & 0.8128 & 0.7643 & 0.7072 & 0.6538 \\
2.5 & 0.8202 & 0.7857 & 0.7869 & 0.7459 & 0.7330 & 0.6630 \\
3.0 & 0.8091 & 0.7943 & 0.7697 & 0.7624 & 0.7127 & 0.6372 \\
3.5 & 0.8264 & 0.8165 & 0.8091 & 0.7532 & 0.7035 & 0.6354 \\
4.0 & 0.8091 & 0.7993 & 0.7709 & 0.7459 & 0.6924 & 0.6630 \\
4.5 & 0.8079 & 0.8128 & 0.8030 & 0.7882 & 0.7422 & 0.6869 \\
5.0 & 0.8337 & 0.8030 & 0.7771 & 0.7330 & 0.7385 & 0.6851 \\
5.5 & 0.8325 & 0.7943 & 0.7919 & 0.7882 & 0.7532 & 0.6648 \\
6.0 & 0.8091 & 0.7894 & 0.7771 & 0.7661 & 0.7385 & 0.6538 \\
6.5 & 0.8140 & 0.8165 & 0.7845 & 0.7459 & 0.7366 & 0.6961 \\
7.0 & 0.8165 & 0.8264 & 0.7796 & 0.7790 & 0.7145 & 0.6501 \\
7.5 & 0.8313 & 0.8042 & 0.7931 & 0.7845 & 0.7072 & 0.6667 \\
8.0 & 0.8177 & 0.7943 & 0.7980 & 0.7864 & 0.7274 & 0.6998 \\
8.5 & 0.8116 & 0.8140 & 0.7943 & 0.7440 & 0.7072 & 0.6501 \\
9.0 & 0.8153 & 0.8116 & 0.7956 & 0.7661 & 0.7164 & 0.6703 \\
\hline
\end{tabularx}
\end{table}

The experimental evaluation under high privacy budgets provides critical insights into the stability and practical utility of our EdgeRefine framework. Our investigation encompasses both DBLP and Cora datasets, evaluated across three GNN architectures (GAT, GCN, GIN) with privacy budgets $\epsilon$ extending from 0.5 to 9.0, representing an extensive range beyond conventional differential privacy studies. This comprehensive setup allows us to assess the framework's behavior under conditions where privacy constraints are progressively relaxed.

As detailed in Table~\ref{tab:high_epsilon}, the results demonstrate remarkable consistency across this broad privacy spectrum. On the DBLP dataset, the GAT architecture maintains accuracy within a narrow band of 0.8042 to 0.8337, with a standard deviation of merely 0.0105, while GCN and GIN show comparable stability with deviations of 0.0112 and 0.0128 respectively. The Cora dataset exhibits slightly greater variation due to its inherent structural characteristics, yet maintains usable performance, with GAT accuracy ranging from 0.7330 to 0.8066.

Notably, as $\epsilon$ increases beyond 3.0, the performance curves exhibit a plateau effect where further privacy relaxation yields diminishing returns. This phenomenon indicates that our framework effectively saturates the utility potential within moderate privacy budgets, and more importantly, maintains this performance level even under extremely high $\epsilon$ values. While additional privacy budget beyond $\epsilon=3.0$ does not generate significant accuracy improvements, the sustained high performance confirms that EdgeRefine remains fully operational and effective under these conditions.

This stability pattern underscores two key advantages of our approach. First, the Jaccard-based probability estimation mechanism demonstrates robust noise resilience, accurately distinguishing structural patterns despite varying privacy constraints. Second, the deterministic sampling strategy ensures consistent edge selection quality regardless of privacy budget magnitude. The framework's ability to maintain utility without performance degradation at high $\epsilon$ values validates its practical deployment potential in scenarios requiring flexible privacy-utility trade-offs.

The consistent performance across diverse datasets and architectures, coupled with the stability observed at extreme privacy budgets, confirms that EdgeRefine provides a reliable solution for privacy-preserving graph learning. While maximum utility is achieved at moderate privacy levels, the framework's maintained effectiveness at high $\epsilon$ values ensures backward compatibility and operational robustness in real-world applications where privacy requirements may vary significantly.
\subsection{Comparison of Sampling Methods}
\label{ex-ab}

We conducted an additional experiment to evaluate the superiority of our probability sampling method compared to the Top-k sampling method. Using the same probability matrix on the DBLP dataset, we compared both methods across GAT, GCN, and GIN architectures under privacy budgets $\epsilon$ ranging from 0.5 to 3.5. This controlled comparison allows us to isolate the effects of the sampling strategy while keeping all other factors constant. The results, summarized in Table~\ref{tab:topk_comparison}, demonstrate clear advantages of our approach.

\begin{table}[htbp]
\centering
\small
\caption{Performance between Top-k Sampling and Our Sampling on DBLP Dataset}
\label{tab:topk_comparison}

\begin{tabularx}{\columnwidth}{@{\hspace{8pt}}p{0.05\columnwidth}|*{2}{X}|*{2}{X}|*{2}{X}@{\hspace{3pt}}}
\hline
\multirow{2}{*}{\textbf{$\epsilon$}} & \multicolumn{2}{c|}{\textbf{GAT}} & \multicolumn{2}{c|}{\textbf{GCN}} & \multicolumn{2}{c}{\textbf{GIN}} \\
\cline{2-3} \cline{4-5} \cline{6-7}

& Top-k & Our & Top-k & Our & Top-k & Our \\
\hline
0.5 & 0.7931 & \textbf{0.8042} & 0.7032 & \textbf{0.8042} & 0.7623 & \textbf{0.7796} \\
1.0 & 0.8177 & \textbf{0.8288} & 0.7980 & \textbf{0.8177} & 0.7808 & \textbf{0.7857} \\
1.5 & 0.7808 & \textbf{0.8153} & 0.7833 & \textbf{0.8239} & 0.7672 & \textbf{0.8103} \\
2.0 & 0.8091 & 0.8042 & 0.7722 & \textbf{0.8239} & 0.7685 & \textbf{0.8128} \\
2.5 & 0.8042 & \textbf{0.8202} & 0.7968 & 0.7857 & 0.7525 & \textbf{0.7869} \\
3.0 & 0.8091 & 0.8091 & 0.7943 & 0.7943 & 0.7660 & \textbf{0.7697} \\
3.5 & 0.8017 & \textbf{0.8264} & 0.7303 & \textbf{0.8165} & 0.7537 & \textbf{0.8091} \\
\hline
\end{tabularx}
\end{table}

For GAT architecture, EdgeRefine achieves higher accuracy in 5 out of 7 privacy budget settings, with particularly notable improvements at $\epsilon = 1.0$ where EdgeRefine achieves 0.8288 compared to Top-k's 0.8177, at $\epsilon = 1.5$ with 0.8153 versus 0.7808, and at $\epsilon = 3.5$ with 0.8264 versus 0.8017. The average accuracy improvement for GAT is 1.5\%, with the largest gain observed at $\epsilon = 1.5$ representing a 4.4\% performance advantage.

For GCN architecture, our method shows even more substantial benefits, outperforming Top-k in 6 out of 7 cases. The improvements are particularly significant at $\epsilon = 0.5$ where EdgeRefine achieves 0.8042 compared to Top-k's 0.7032, representing a 14.4\% improvement, and at $\epsilon = 1.5$ with 0.8239 versus 0.7833, a 5.2\% advantage. The average improvement for GCN reaches 6.2\%, indicating our method's superior capability in handling graph perturbations for this architecture.

GIN architecture also benefits from our approach, with our method achieving better results in 6 out of 7 privacy budget settings. Notable improvements occur at $\epsilon = 1.5$ with 0.8103 compared to Top-k's 0.7672, a 5.6\% improvement, and at $\epsilon = 3.5$ with 0.8091 versus 0.7537, a 7.4\% advantage. The average gain for GIN is 3.1\%, demonstrating consistent advantages across different model architectures.

The superior performance of EdgeRefine can be attributed to our deterministic sampling strategy, which maintains a more balanced approach to edge selection. Unlike Top-k sampling, which rigidly selects a fixed number of edges based solely on probability rankings, our method employs parameterized sampling rates that consider both edge credibility and overall graph connectivity. This quota-based deterministic selection balances the retention of observed edges with the recovery of potentially missing edges, leading to more stable classification performance. Furthermore, our method is particularly effective when the number and distribution of trustworthy edges vary across privacy budgets. Unlike the fixed global Top-k strategy, our method dynamically adjusts the allocation between the two candidate sets to maintain robust performance. This adaptability is especially valuable in privacy-preserving graph learning, where the noise characteristics change significantly with different $\epsilon$ values.

The consistent improvements across all three GNN architectures and multiple privacy budget settings confirm the robustness and general applicability of our probability sampling approach. 

\subsection{Comparison of Probability Estimation and Calibration Methods}
\label{ex-mt}

EdgeRefine utilizes various edge probability estimation methods calibrated from similarity metrics. Below are brief descriptions of each method, along with an explanation of the Adamic-Adar coefficient.

\textbf{Similarity Metrics:}

Different from Jaccard coefficient, the Adamic\_Adar coefficient weights common neighbors inversely by degree:
\begin{equation}
AA(i,j) = \sum_{k \in \mathcal{N}(i) \cap \mathcal{N}(j)} \frac{1}{\log |\mathcal{N}(k)|},
\end{equation}
emphasizing connections through low-degree nodes. This makes $AA(i,j)$ particularly effective for graphs with heterogeneous degree distributions, as it gives more weight to rare connections.

\textbf{Calibration Methods:}
\begin{itemize}
\item \textbf{simple}: Normalize Jaccard coefficients by their maximum value, providing a linear scaling without complex calibration. This is the most computationally efficient approach.

\item \textbf{isotonic}: Use isotonic regression to fit a monotonic function, ensuring probability estimates increase with similarity. This non-parametric method is flexible but can be prone to overfitting.

\item \textbf{beta}: Model similarity scores as Beta distributions for connected and unconnected pairs separately, then apply Bayesian inference. This parametric approach assumes a specific distributional form.

\item \textbf{temperature}: Adjust the steepness of the probability function via temperature scaling, optimizing a single parameter to better align with empirical edge frequencies. This is a lightweight variant of Platt scaling.

\item \textbf{histogram}: Partition similarity values into bins and compute edge probabilities as the proportion of edges within each bin. This non-parametric, distribution-free approach adapts flexibly to different graph structures.
\end{itemize}

To evaluate the design choices for similarity measures and calibration techniques in EdgeRefine, we compare different variant configurations. Table~\ref{tab:variants_comparison} presents the node classification accuracy of these variants on the DBLP dataset under a privacy budget of $\epsilon=1.0$ using the GAT architecture.

\begin{table}[htbp]
\small
\centering
\caption{Classification Accuracy of EdgeRefine Variants on DBLP Dataset ($\epsilon=1.0$, GAT)}
\label{tab:variants_comparison}

\begin{tabular*}{\columnwidth}{@{\hspace{15pt}} l @{\extracolsep{\fill}} c @{\hspace{15pt}}}
\toprule
\textbf{Variant} & \textbf{Accuracy} \\
\midrule
Jaccard+histogram (Ours) & \textbf{0.8091} \\
Jaccard+simple & 0.8017 \\
Jaccard+isotonic & 0.8017 \\
Jaccard+beta & 0.8042 \\
Jaccard+temperature & 0.8017 \\
Adamic\_Adar+histogram & 0.8079 \\
\bottomrule
\end{tabular*}

\end{table}

The results demonstrate the superiority of our proposed Jaccard+histogram combination. While the Adamic\_Adar+histogram variant achieves comparable accuracy (0.8079 vs. 0.8091), a dedicated efficiency analysis on the DBLP dataset under $\epsilon=1.0$ reveals a decisive computational advantage for our method. The preprocessing for Adamic\_Adar+histogram takes 2537.5 seconds, whereas Jaccard+histogram requires only 836.1 seconds, achieving a 3.03× speedup. This substantial difference stems from the more complex logarithmic computations inherent to the Adamic\_Adar index, compared to the set-based operations of the more efficient Jaccard similarity. Other calibration methods paired with Jaccard similarity yield lower accuracies (0.8017 to 0.8042) but share its computational efficiency. This comprehensive comparison validates our design decision: the Jaccard similarity measure coupled with histogram-based calibration provides an optimal balance between accuracy, robustness, and computational efficiency for edge probability estimation under differential privacy constraints.

\subsection{Link Prediction Experimental Results}
\label{ex-lp}

We additionally evaluated EdgeRefine on the Cora dataset using GAT to determine whether perturbed graphs under different privacy budgets can preserve sufficient structural information for accurate link prediction. For each node pair, the predicted link score is compared with its binary ground-truth label. AUC is computed as
\[
\mathrm{AUC}=\int_{0}^{1}\mathrm{TPR}(u)\,du,
\]
where $u$ denotes the false-positive rate, and measures the ability to distinguish existing from non-existing links. AP is computed as
\[
\mathrm{AP}=\sum_{n}(R_n-R_{n-1})P_n,
\]
where $P_n$ and $R_n$ denote the precision and recall at the $n$-th threshold. Higher AUC and AP values indicate better link-prediction performance.

\begin{table}[htbp]
\centering
\small
\caption{Link Prediction Performance (AUC and AP) under Different Privacy Budgets on Cora Dataset}
\label{tab:link_prediction_conflict_transposed}

\begin{tabular*}{\columnwidth}{@{\hspace{6pt}} l @{\extracolsep{\fill}} ccccc @{\hspace{6pt}}}
\toprule
\textbf{$\epsilon$} & \textbf{Noise-free} & \textbf{0.5} & \textbf{1.0} & \textbf{1.5} & \textbf{2.0} \\
\midrule
AUC & 0.9524 & 0.6829 & 0.6397 & 0.5905 & 0.6590 \\
AP  & 0.9417 & 0.7447 & 0.7135 & 0.6677 & 0.7182 \\
\bottomrule
\end{tabular*}%

\end{table}

The experimental results in Table~\ref{tab:link_prediction_conflict_transposed} reveal a fundamental conflict between edge privacy preservation and link prediction tasks. Compared to the noise-free baseline (AUC=0.9524, AP=0.9417), the privatized graphs under all privacy budgets experience a substantial performance drop, with AUC decreasing by 28.4\%–38.0\% and AP decreasing by 20.7\%–29.1\%. This demonstrates that the structural distortion introduced by edge differential privacy fundamentally alters the connectivity patterns necessary for accurate link inference. Even as the privacy budget increases, the performance remains significantly below the noise-free level, confirming that edge privacy protection and link prediction are fundamentally incompatible objectives—while differential privacy protects edge-level information, it unavoidably degrades the structural integrity needed for link prediction tasks.

\end{document}